\documentclass{article}

\usepackage[final, nonatbib]{neurips_2021}

\usepackage[utf8]{inputenc} 
\usepackage[T1]{fontenc}    
\usepackage{hyperref}       
\hypersetup{colorlinks}

\usepackage{url}            
\usepackage{booktabs}       
\usepackage{amsfonts}       
\usepackage{nicefrac}       
\usepackage{microtype}      
\usepackage{xcolor}         

\usepackage{tikzsymbols}
\usetikzlibrary{arrows}
\usetikzlibrary{chains}
\usetikzlibrary{patterns}
\usepackage{tikz-qtree}
\usetikzlibrary{shapes.multipart}
\usetikzlibrary{positioning}
\usetikzlibrary{arrows.meta}
\usepackage{pgfplots}

\usetikzlibrary{calc}
\usetikzlibrary{positioning}
\usepgflibrary{fpu}
\usepgfplotslibrary{fillbetween}

\usetikzlibrary{external}
\pgfplotsset{compat=newest}

\usepackage{amsmath}
\usepackage{amssymb}
\usepackage{placeins}
\usepackage{subcaption}

\newenvironment{customlegend}[1][]{%
    \begingroup
    \csname pgfplots@init@cleared@structures\endcsname
    \pgfplotsset{#1}%
}{%
    \csname pgfplots@createlegend\endcsname
    \endgroup
}%
\def\addlegendimage{\csname pgfplots@addlegendimage\endcsname}

\title{Exploring the Limits of Epistemic Uncertainty Quantification in Low-Shot Settings}

\author{%
    \parbox{\linewidth}{
        Matias Valdenegro-Toro\\
    }
    ~\\
    \parbox{\linewidth}{
        German Research Center for Artificial Intelligence, 28359 Bremen, Germany.\\
        ~\\        
        \texttt{matias.valdenegro@dfki.de}
    }
}

\begin{document}

\maketitle

\begin{abstract}
    Uncertainty quantification in neural network promises to increase safety of AI systems, but it is not clear how performance might vary with the training set size. In this paper we evaluate seven uncertainty methods on Fashion MNIST and CIFAR10, as we sub-sample and produce varied training set sizes. We find that calibration error and out of distribution detection performance strongly depend on the training set size, with most methods being miscalibrated on the test set with small training sets. Gradient-based methods seem to poorly estimate epistemic uncertainty and are the most affected by training set size. We expect our results can guide future research into uncertainty quantification and help practitioners select methods based on their particular available data.
\end{abstract}

\section{Introduction}

Neural networks are now ubiquitous for many tasks in various fields like computer vision \cite{kendall2017uncertainties}, natural language processing, and autonomous driving \cite{feng2021review} \cite{sunderhauf2018limits}. Despite their success, these methods generally have problems quantifying their own uncertainty \cite{guo2017calibration}, which is necessary for safety, reliability, and overall trustworthiness, particularly when used in human environments.

Bayesian Deep Learning promises good uncertainty estimates \cite{wilson2020case}, but methods often rely in approximations to the bayesian posterior, or quantify uncertainty in approximate \cite{wen2018flipout} \cite{blundell2015weight} or non-bayesian ways \cite{lakshminarayanan2017simple}. Evaluating the quality of output uncertainty is difficult as there are no labels. For specific details we refer the reader to \cite{gawlikowski2021survey}.

Real-world datasets have multiple issues that are not present in academic benchmarks (like CIFAR10, Fashion MNIST, ImageNet, etc), such as low number of samples. There is clear interest on information how uncertainty methods perform under low show scenarios, where the training set might have little number of samples (less than 1000). It is unclear how these methods might perform, and there is a relationship with model (also known as epistemic) uncertainty, where the lack of information in the training set produces increased uncertainty at the output \cite{der2009aleatory} \cite{hullermeier2021aleatoric}.

\begin{figure}[!ht]
    \centering
    \includegraphics[width=0.87\textwidth]{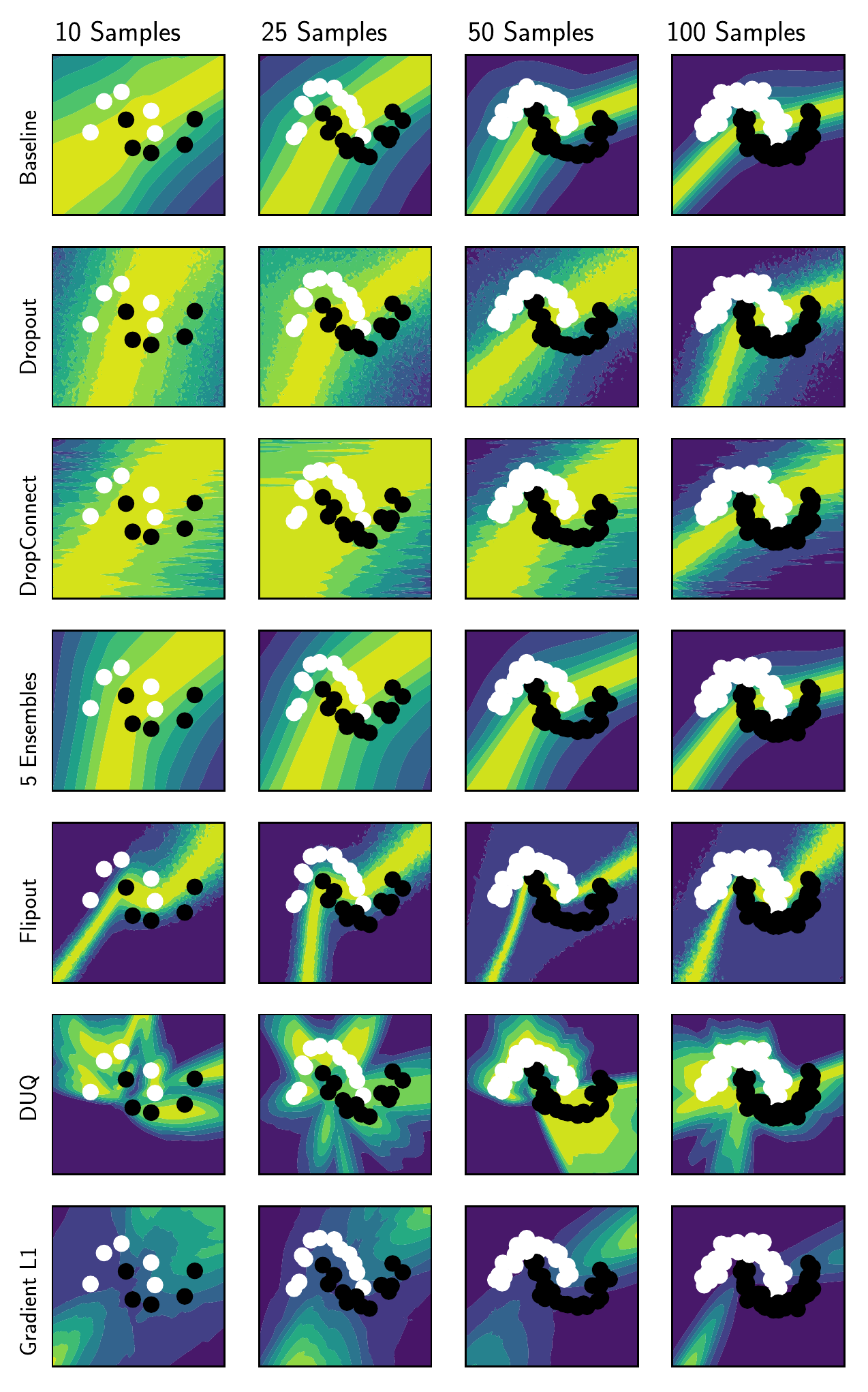}
    \caption{Comparison of uncertainty/confidence in the Two Moons dataset across multiple methods and training set samples per class. This is a synthetic dataset and is available in scikit-learn \cite{pedregosa2011scikit} at \url{https://scikit-learn.org/stable/modules/generated/sklearn.datasets.make_moons.html}. This example visually supports our main conclusion that most uncertainty methods produce miscalibrated predictions at small training set sizes, and that gradient uncertainty is counter-intuitive and overconfident.}
    \label{two_moons_example}
\end{figure}

In this paper, we evaluate multiple uncertainty methods on sub-sampled versions of two toy datasets, and the Fashion MNIST and CIFAR10 training sets, over a variety of metrics for uncertainty quantification, including entropy, maximum probability, calibration error, and out of distribution performance over several combinations of datasets. An example of our evaluation on the Two Moons dataset (classification) is shown in Figure \ref{two_moons_example} and a toy example (regression) is shown in Figure \ref{toy_reg_example}, where it is clear that uncertainty varies considerably with the training set size.

\section{Experimental Setup}

Our principal setup is to sub-sample the training set and train a neural network on this dataset, evaluating a set of metrics related to uncertainty. We sub-sample training sets to a fixed number of samples per class (SPC), namely $S \in [1, 5, 10, 50, 100, 250, 500, 1000, 5000]$ Sub-sampling happens by randomly drawing a fixed number of samples for each class, without replacement. For each value of $s$, we perform $K = 5$ trials where one model is trained on the sub-sampled training set, and metrics are evaluated on each trial. We report the mean and standard deviation of each metric across the $K$ trials.

\textbf{Uncertainty Methods}. We evaluate MC-Dropout (DO) \cite{gal2016dropout}, MC-DropConnect (DC) \cite{mobiny2021dropconnect}, Deep Ensembles (DE) \cite{lakshminarayanan2017simple}, Direct Uncertainty Quantification (DUQ) \cite{van2020uncertainty}, Variational Inference with Flipout (VI) \cite{wen2018flipout}, and Gradient-based uncertainty (GD) \cite{oberdiek2018classification}. This selection covers scalable as well as approximate methods and recent advances. Detailed descriptions and hyper-parameters of each method is available in the appendix. We also use a standard CNN architecture without any uncertainty quantification as a comparison baseline (denoted as BL).

\textbf{Metrics}. We use a selection of metrics that measure different aspects of uncertainty quality. As basic metrics we evaluate accuracy, entropy, and maximum probability, all in the test set. For entropy and maximum probability, we are interested in measuring the confidence level of the classifier, and how it changes with the size of the training set. Some additional metrics are:

\begin{itemize}
    \item \textbf{Expected Calibration Error}. We evaluate expected calibration error \cite{guo2017calibration} in the train and test sets. We expect that classifiers should be calibrated independent of the training set size.
    \item \textbf{Out of Distribution Detection}. For each dataset, we define an OOD dataset, and then evaluate the area under the ROC curve (AUC) for different combinations of in distribution (ID) and out of distribution  (OOD) datasets, based both on the predictive entropy and maximum probability. The dataset combinations are:
    \begin{itemize}
        \item Test ID vs Test OOD. This is the standard OOD benchmark that is reported many times in the literature \cite{lakshminarayanan2017simple} \cite{hendrycks2017baseline}.
        \item Training ID vs Test OOD. This setting evaluates information in the training set uncertainty, as this dataset increases in size, against the out of distribution test set. This is similar to the below experiment but without evaluating for generalization that is usually done with ID test sets.
        \item Training ID vs Test ID. Here we evaluate discrimination between the train and test sets in the in-distribution setting. We expect that as the training set size increases, the AUC should decrease and settle around 0.5, as the uncertainty of the train and test sets would be very similar. This setup validates this assumption experimentally.
    \end{itemize}
    Note that for gradient uncertainty methods, only maximum probability is available, as only a single confidence value is produced.
\end{itemize}

\textbf{Datasets}. We use Fashion MNIST (with MNIST as OOD dataset), and CIFAR10 (with SVHN as OOD dataset).

\textbf{Training}. We train each model for 100 epochs using Adam, with a batch size $B = 64$. All models converge in this setting at all training set sizes. A categorical cross-entropy loss is used for all models except for DUQ, which uses a binary cross-entropy loss for training. Information about the network architecture is available in the appendix.


\section{Experimental Evaluation and Results}

\newcommand{\plotwitherrorareas}[5]{
    
    \addplot+[mark = none, color=#2] table[x  = spc, y  = #4, col sep = semicolon] {#1};
    \addlegendentry{#3}
    
    \addplot [name path=upper, draw=none, forget plot] table[x=spc, y expr=\thisrow{#4} + 0.5 * \thisrow{#5}, col sep = semicolon] {#1};
    \addplot [name path=lower, draw=none, forget plot] table[x=spc, y expr=\thisrow{#4} - 0.5 * \thisrow{#5}, col sep = semicolon] {#1};
    \addplot [fill=#2!10, forget plot] fill between[of=upper and lower];   
}

\begin{figure*}[!htb]
    \centering
    \begin{tikzpicture}
        \begin{customlegend}[legend columns = 7,legend style = {column sep=1ex}, legend cell align = left,
            legend entries={DE, DO, DC, DUQ, BL, VI, GD}]
            \addlegendimage{mark=none,blue, only marks}
            \addlegendimage{mark=none,red, only marks}
            \addlegendimage{mark=none,green, only marks}
            \addlegendimage{mark=none,black, only marks}
            \addlegendimage{mark=none,magenta, only marks}
            \addlegendimage{mark=none,gray, only marks}
            \addlegendimage{mark=none,orange, only marks}
        \end{customlegend}
    \end{tikzpicture}
    
    \begin{subfigure}{0.49\textwidth}
        \begin{tikzpicture}
            \begin{axis}[height = 0.2 \textheight, width = \textwidth, xlabel={Samples per Class}, ylabel={Test Accuracy}, ymajorgrids=false, xmajorgrids=false, grid style=dashed, legend pos = south east, legend style={font=\scriptsize}, tick label style={font=\tiny,rotate=40}, xtick=data, xmode=log, log ticks with fixed point]
                
                \plotwitherrorareas{data/entropy-vs-SPC-fashion_mnist-results-miniVGG-deepensemble-combined.csv}{blue}{DE}{mean_acc}{std_acc}
                \plotwitherrorareas{data/entropy-vs-SPC-fashion_mnist-results-miniVGG-dropout-combined.csv}{red}{DO}{mean_acc}{std_acc}
                \plotwitherrorareas{data/entropy-vs-SPC-fashion_mnist-results-miniVGG-dropconnect-combined.csv}{green}{DC}{mean_acc}{std_acc}
                \plotwitherrorareas{data/entropy-vs-SPC-fashion_mnist-results-miniVGG-duq-combined.csv}{black}{DUQ}{mean_acc}{std_acc}
                \plotwitherrorareas{data/entropy-vs-SPC-fashion_mnist-results-miniVGG-baseline-combined.csv}{magenta}{BL}{mean_acc}{std_acc}
                \plotwitherrorareas{data/entropy-vs-SPC-fashion_mnist-results-miniVGG-flipout-combined.csv}{gray}{VI}{mean_acc}{std_acc}
                \legend{}
            \end{axis}		            
        \end{tikzpicture}
        \caption{Accuracy}
    \end{subfigure}
    \begin{subfigure}{0.49\textwidth}
        \begin{tikzpicture}
            \begin{axis}[height = 0.2 \textheight, width = \textwidth, xlabel={Samples per Class}, ylabel={Entropy}, ymajorgrids=false, xmajorgrids=false, grid style=dashed, legend pos = north east, legend style={font=\scriptsize}, tick label style={font=\tiny,rotate=40}, xtick=data, xmode=log, log ticks with fixed point]
                
                \plotwitherrorareas{data/entropy-vs-SPC-fashion_mnist-results-miniVGG-deepensemble-combined.csv}{blue}{DE}{mean_mean_entropy}{std_mean_entropy}
                \plotwitherrorareas{data/entropy-vs-SPC-fashion_mnist-results-miniVGG-dropout-combined.csv}{red}{DO}{mean_mean_entropy}{std_mean_entropy}
                \plotwitherrorareas{data/entropy-vs-SPC-fashion_mnist-results-miniVGG-dropconnect-combined.csv}{green}{DC}{mean_mean_entropy}{std_mean_entropy}
                \plotwitherrorareas{data/entropy-vs-SPC-fashion_mnist-results-miniVGG-duq-combined.csv}{black}{DUQ}{mean_mean_entropy}{std_mean_entropy}
                \plotwitherrorareas{data/entropy-vs-SPC-fashion_mnist-results-miniVGG-baseline-combined.csv}{magenta}{BL}{mean_mean_entropy}{std_mean_entropy}
                \plotwitherrorareas{data/entropy-vs-SPC-fashion_mnist-results-miniVGG-flipout-combined.csv}{gray}{VI}{mean_mean_entropy}{std_mean_entropy}
                \legend{}
            \end{axis}		            
        \end{tikzpicture}
        \caption{Entropy}
    \end{subfigure}
    
    \begin{subfigure}{0.49\textwidth}
        \begin{tikzpicture}
            \begin{axis}[height = 0.2 \textheight, width = \textwidth, xlabel={Samples per Class}, ylabel={Max Prob}, ymajorgrids=false, xmajorgrids=false, grid style=dashed, legend pos = south east, legend style={font=\scriptsize}, tick label style={font=\tiny,rotate=40}, xtick=data, xmode=log, log ticks with fixed point]
                
                \plotwitherrorareas{data/entropy-vs-SPC-fashion_mnist-results-miniVGG-deepensemble-combined.csv}{blue}{DE}{mean_mean_maxprob}{std_mean_maxprob}
                \plotwitherrorareas{data/entropy-vs-SPC-fashion_mnist-results-miniVGG-dropout-combined.csv}{red}{DO}{mean_mean_maxprob}{std_mean_maxprob}
                \plotwitherrorareas{data/entropy-vs-SPC-fashion_mnist-results-miniVGG-dropconnect-combined.csv}{green}{DC}{mean_mean_maxprob}{std_mean_maxprob}
                \plotwitherrorareas{data/entropy-vs-SPC-fashion_mnist-results-miniVGG-duq-combined.csv}{black}{DUQ}{mean_mean_maxprob}{std_mean_maxprob}
                \plotwitherrorareas{data/entropy-vs-SPC-fashion_mnist-results-miniVGG-baseline-combined.csv}{magenta}{BL}{mean_mean_maxprob}{std_mean_maxprob}
                \plotwitherrorareas{data/entropy-vs-SPC-fashion_mnist-results-miniVGG-flipout-combined.csv}{gray}{VI}{mean_mean_maxprob}{std_mean_maxprob}
                \plotwitherrorareas{data/maxprob-vs-SPC-fashion_mnist-results-miniVGG-gradient-l1_norm-combined.csv}{orange}{GD}{mean_mean_maxprob}{std_mean_maxprob}
                \legend{}
            \end{axis}		            
        \end{tikzpicture}
        \caption{Max Prob}
    \end{subfigure}
    
    \begin{subfigure}{0.49\textwidth}
        \begin{tikzpicture}
            \begin{axis}[height = 0.2 \textheight, width = \textwidth, xlabel={Samples per Class}, ylabel={ECE}, ymajorgrids=false, xmajorgrids=false, grid style=dashed, legend pos = north east, legend style={font=\scriptsize}, tick label style={font=\tiny,rotate=40}, xtick=data, xmode=log, log ticks with fixed point, ymin=0.0, ymax=0.4]
                
                \plotwitherrorareas{data/entropy-vs-SPC-fashion_mnist-results-miniVGG-deepensemble-combined.csv}{blue}{DE}{mean_train_ece}{std_train_ece}
                \plotwitherrorareas{data/entropy-vs-SPC-fashion_mnist-results-miniVGG-dropout-combined.csv}{red}{DO}{mean_train_ece}{std_train_ece}
                \plotwitherrorareas{data/entropy-vs-SPC-fashion_mnist-results-miniVGG-dropconnect-combined.csv}{green}{DC}{mean_train_ece}{std_train_ece}
                \plotwitherrorareas{data/entropy-vs-SPC-fashion_mnist-results-miniVGG-duq-combined.csv}{black}{DUQ}{mean_train_ece}{std_train_ece}
                \plotwitherrorareas{data/entropy-vs-SPC-fashion_mnist-results-miniVGG-baseline-combined.csv}{magenta}{BL}{mean_train_ece}{std_train_ece}
                \plotwitherrorareas{data/entropy-vs-SPC-fashion_mnist-results-miniVGG-flipout-combined.csv}{gray}{VI}{mean_train_ece}{std_train_ece}
                \plotwitherrorareas{data/maxprob-vs-SPC-fashion_mnist-results-miniVGG-gradient-l1_norm-combined.csv}{orange}{GD}{mean_train_ece}{std_train_ece}
                \legend{}
            \end{axis}		            
        \end{tikzpicture}
        \caption{Train ECE}
    \end{subfigure}
    \begin{subfigure}{0.49\textwidth}
        \begin{tikzpicture}
            \begin{axis}[height = 0.2 \textheight, width = \textwidth, xlabel={Samples per Class}, ylabel={ECE}, ymajorgrids=false, xmajorgrids=false, grid style=dashed, legend pos = north east, legend style={font=\scriptsize}, tick label style={font=\tiny,rotate=40}, xtick=data, xmode=log, log ticks with fixed point, ymin=0.0, ymax=0.4]
                
                \plotwitherrorareas{data/entropy-vs-SPC-fashion_mnist-results-miniVGG-deepensemble-combined.csv}{blue}{DE}{mean_ece}{std_ece}
                \plotwitherrorareas{data/entropy-vs-SPC-fashion_mnist-results-miniVGG-dropout-combined.csv}{red}{DO}{mean_ece}{std_ece}
                \plotwitherrorareas{data/entropy-vs-SPC-fashion_mnist-results-miniVGG-dropconnect-combined.csv}{green}{DC}{mean_ece}{std_ece}
                \plotwitherrorareas{data/entropy-vs-SPC-fashion_mnist-results-miniVGG-duq-combined.csv}{black}{DUQ}{mean_ece}{std_ece}
                \plotwitherrorareas{data/entropy-vs-SPC-fashion_mnist-results-miniVGG-baseline-combined.csv}{magenta}{BL}{mean_ece}{std_ece}
                \plotwitherrorareas{data/entropy-vs-SPC-fashion_mnist-results-miniVGG-flipout-combined.csv}{gray}{VI}{mean_ece}{std_ece}
                \plotwitherrorareas{data/maxprob-vs-SPC-fashion_mnist-results-miniVGG-gradient-l1_norm-combined.csv}{orange}{GD}{mean_ece}{std_ece}
                \legend{}
            \end{axis}		            
        \end{tikzpicture}
        \caption{Test ECE}
    \end{subfigure}
    
    \begin{subfigure}{0.49\textwidth}
        \begin{tikzpicture}
            \begin{axis}[height = 0.2 \textheight, width = \textwidth, xlabel={Samples per Class}, ylabel={OOD AUC}, ymajorgrids=false, xmajorgrids=false, grid style=dashed, legend pos = south east, legend style={font=\scriptsize}, tick label style={font=\tiny,rotate=40}, xtick=data, xmode=log, log ticks with fixed point, ymin=0.2, ymax=1.0]
                
                \plotwitherrorareas{data/entropy-vs-SPC-fashion_mnist-results-miniVGG-deepensemble-combined.csv}{blue}{DE}{mean_ood_auc_entropy}{std_ood_auc_entropy}  
                \plotwitherrorareas{data/entropy-vs-SPC-fashion_mnist-results-miniVGG-dropout-combined.csv}{red}{DO}{mean_ood_auc_entropy}{std_ood_auc_entropy}
                \plotwitherrorareas{data/entropy-vs-SPC-fashion_mnist-results-miniVGG-dropconnect-combined.csv}{green}{DC}{mean_ood_auc_entropy}{std_ood_auc_entropy}
                \plotwitherrorareas{data/entropy-vs-SPC-fashion_mnist-results-miniVGG-duq-combined.csv}{black}{DUQ}{mean_ood_auc_entropy}{std_ood_auc_entropy}
                \plotwitherrorareas{data/entropy-vs-SPC-fashion_mnist-results-miniVGG-baseline-combined.csv}{magenta}{BL}{mean_ood_auc_entropy}{std_ood_auc_entropy}
                \plotwitherrorareas{data/entropy-vs-SPC-fashion_mnist-results-miniVGG-flipout-combined.csv}{gray}{VI}{mean_ood_auc_entropy}{std_ood_auc_entropy}                
                \legend{}
            \end{axis}		            
        \end{tikzpicture}
        \caption{Test/OOD AUC Entropy}
    \end{subfigure}
    \begin{subfigure}{0.49\textwidth}
        \begin{tikzpicture}
            \begin{axis}[height = 0.2 \textheight, width = \textwidth, xlabel={Samples per Class}, ylabel={OOD AUC}, ymajorgrids=false, xmajorgrids=false, grid style=dashed, tick label style={font=\tiny,rotate=40}, xtick=data, xmode=log, log ticks with fixed point, ymin=0.2, ymax=1.0]
                
                \plotwitherrorareas{data/entropy-vs-SPC-fashion_mnist-results-miniVGG-deepensemble-combined.csv}{blue}{DE}{mean_ood_auc_maxprob}{std_ood_auc_maxprob}
                \plotwitherrorareas{data/entropy-vs-SPC-fashion_mnist-results-miniVGG-dropout-combined.csv}{red}{DO}{mean_ood_auc_maxprob}{std_ood_auc_maxprob}
                \plotwitherrorareas{data/entropy-vs-SPC-fashion_mnist-results-miniVGG-dropconnect-combined.csv}{green}{DC}{mean_ood_auc_maxprob}{std_ood_auc_maxprob}
                \plotwitherrorareas{data/entropy-vs-SPC-fashion_mnist-results-miniVGG-duq-combined.csv}{black}{DUQ}{mean_ood_auc_maxprob}{std_ood_auc_maxprob}
                \plotwitherrorareas{data/entropy-vs-SPC-fashion_mnist-results-miniVGG-baseline-combined.csv}{magenta}{BL}{mean_ood_auc_maxprob}{std_ood_auc_maxprob}
                \plotwitherrorareas{data/entropy-vs-SPC-fashion_mnist-results-miniVGG-flipout-combined.csv}{gray}{VI}{mean_ood_auc_maxprob}{std_ood_auc_maxprob}
                \plotwitherrorareas{data/maxprob-vs-SPC-fashion_mnist-results-miniVGG-gradient-l1_norm-combined.csv}{orange}{GD}{mean_ood_auc_maxprob}{std_ood_auc_maxprob}
                \legend{}
            \end{axis}		            
        \end{tikzpicture}
        \caption{Test/OOD AUC Max Prob}
    \end{subfigure}
    
    \caption{Comparison of uncertainty as size of the training set is varied on Fashion MNIST.}
    \label{comparison_plots_fmnist}
\end{figure*}
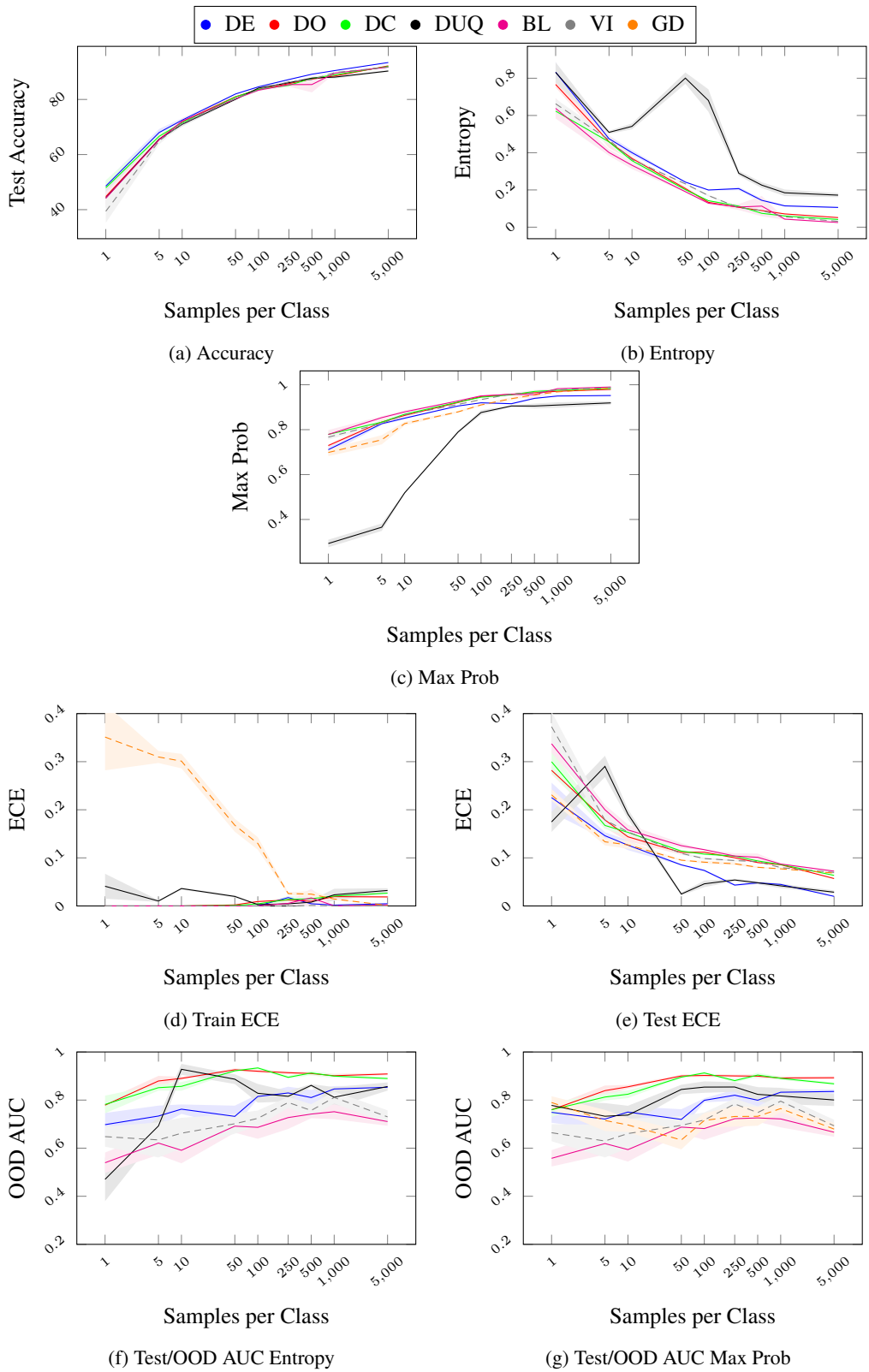

\begin{figure*}[!htb]
    \centering
    \begin{tikzpicture}
        \begin{customlegend}[legend columns = 7,legend style = {column sep=1ex}, legend cell align = left,
            legend entries={DE, DO, DC, DUQ, BL, VI, GD}]
            \addlegendimage{mark=none,blue, only marks}
            \addlegendimage{mark=none,red, only marks}
            \addlegendimage{mark=none,green, only marks}
            \addlegendimage{mark=none,black, only marks}
            \addlegendimage{mark=none,magenta, only marks}
            \addlegendimage{mark=none,gray, only marks}
            \addlegendimage{mark=none,orange, only marks}
        \end{customlegend}
    \end{tikzpicture}
    
    \begin{subfigure}{0.49\textwidth}
        \begin{tikzpicture}
            \begin{axis}[height = 0.2 \textheight, width = \textwidth, xlabel={Samples per Class}, ylabel={Test Accuracy}, ymajorgrids=false, xmajorgrids=false, grid style=dashed, legend pos = south east, legend style={font=\scriptsize}, tick label style={font=\tiny,rotate=40}, xtick=data, xmode=log, log ticks with fixed point]
                
                \plotwitherrorareas{data/entropy-vs-SPC-cifar10-results-miniVGG-deepensemble-combined.csv}{blue}{DE}{mean_acc}{std_acc}
                \plotwitherrorareas{data/entropy-vs-SPC-cifar10-results-miniVGG-dropout-combined.csv}{red}{DO}{mean_acc}{std_acc}
                \plotwitherrorareas{data/entropy-vs-SPC-cifar10-results-miniVGG-dropconnect-combined.csv}{green}{DC}{mean_acc}{std_acc}
                \plotwitherrorareas{data/entropy-vs-SPC-cifar10-results-miniVGG-duq-combined.csv}{black}{DUQ}{mean_acc}{std_acc}
                \plotwitherrorareas{data/entropy-vs-SPC-cifar10-results-miniVGG-baseline-combined.csv}{magenta}{BL}{mean_acc}{std_acc}
                \plotwitherrorareas{data/entropy-vs-SPC-cifar10-results-miniVGG-flipout-combined.csv}{gray}{VI}{mean_acc}{std_acc}
                \legend{}
            \end{axis}		            
        \end{tikzpicture}
        \caption{Accuracy}
    \end{subfigure}
    \begin{subfigure}{0.49\textwidth}
        \begin{tikzpicture}
            \begin{axis}[height = 0.2 \textheight, width = \textwidth, xlabel={Samples per Class}, ylabel={Entropy}, ymajorgrids=false, xmajorgrids=false, grid style=dashed, legend pos = north east, legend style={font=\scriptsize}, tick label style={font=\tiny,rotate=40}, xtick=data, xmode=log, log ticks with fixed point]
                
                \plotwitherrorareas{data/entropy-vs-SPC-cifar10-results-miniVGG-deepensemble-combined.csv}{blue}{DE}{mean_mean_entropy}{std_mean_entropy}
                \plotwitherrorareas{data/entropy-vs-SPC-cifar10-results-miniVGG-dropout-combined.csv}{red}{DO}{mean_mean_entropy}{std_mean_entropy}
                \plotwitherrorareas{data/entropy-vs-SPC-cifar10-results-miniVGG-dropconnect-combined.csv}{green}{DC}{mean_mean_entropy}{std_mean_entropy}
                \plotwitherrorareas{data/entropy-vs-SPC-cifar10-results-miniVGG-duq-combined.csv}{black}{DUQ}{mean_mean_entropy}{std_mean_entropy}
                \plotwitherrorareas{data/entropy-vs-SPC-cifar10-results-miniVGG-baseline-combined.csv}{magenta}{BL}{mean_mean_entropy}{std_mean_entropy}
                \plotwitherrorareas{data/entropy-vs-SPC-cifar10-results-miniVGG-flipout-combined.csv}{gray}{VI}{mean_mean_entropy}{std_mean_entropy}
                \legend{}
            \end{axis}		            
        \end{tikzpicture}
        \caption{Entropy}
    \end{subfigure}
    
    \begin{subfigure}{0.49\textwidth}
        \begin{tikzpicture}
            \begin{axis}[height = 0.2 \textheight, width = \textwidth, xlabel={Samples per Class}, ylabel={Max Prob}, ymajorgrids=false, xmajorgrids=false, grid style=dashed, legend pos = south east, legend style={font=\scriptsize}, tick label style={font=\tiny,rotate=40}, xtick=data, xmode=log, log ticks with fixed point]
                
                \plotwitherrorareas{data/entropy-vs-SPC-cifar10-results-miniVGG-deepensemble-combined.csv}{blue}{DE}{mean_mean_maxprob}{std_mean_maxprob}
                \plotwitherrorareas{data/entropy-vs-SPC-cifar10-results-miniVGG-dropout-combined.csv}{red}{DO}{mean_mean_maxprob}{std_mean_maxprob}
                \plotwitherrorareas{data/entropy-vs-SPC-cifar10-results-miniVGG-dropconnect-combined.csv}{green}{DC}{mean_mean_maxprob}{std_mean_maxprob}
                \plotwitherrorareas{data/entropy-vs-SPC-cifar10-results-miniVGG-duq-combined.csv}{black}{DUQ}{mean_mean_maxprob}{std_mean_maxprob}
                \plotwitherrorareas{data/entropy-vs-SPC-cifar10-results-miniVGG-baseline-combined.csv}{magenta}{BL}{mean_mean_maxprob}{std_mean_maxprob}
                \plotwitherrorareas{data/entropy-vs-SPC-cifar10-results-miniVGG-flipout-combined.csv}{gray}{VI}{mean_mean_maxprob}{std_mean_maxprob}
                \plotwitherrorareas{data/maxprob-vs-SPC-cifar10-results-miniVGG-gradient-l1_norm-combined.csv}{orange}{GD}{mean_mean_maxprob}{std_mean_maxprob}
                \legend{}
            \end{axis}		            
        \end{tikzpicture}
        \caption{Max Prob}
    \end{subfigure}
    
    \begin{subfigure}{0.49\textwidth}
        \begin{tikzpicture}
            \begin{axis}[height = 0.2 \textheight, width = \textwidth, xlabel={Samples per Class}, ylabel={ECE}, ymajorgrids=false, xmajorgrids=false, grid style=dashed, legend pos = north east, legend style={font=\scriptsize}, tick label style={font=\tiny,rotate=40}, xtick=data, xmode=log, log ticks with fixed point, ymin=0.0, ymax=0.6]
                
                \plotwitherrorareas{data/entropy-vs-SPC-cifar10-results-miniVGG-deepensemble-combined.csv}{blue}{DE}{mean_train_ece}{std_train_ece}
                \plotwitherrorareas{data/entropy-vs-SPC-cifar10-results-miniVGG-dropout-combined.csv}{red}{DO}{mean_train_ece}{std_train_ece}
                \plotwitherrorareas{data/entropy-vs-SPC-cifar10-results-miniVGG-dropconnect-combined.csv}{green}{DC}{mean_train_ece}{std_train_ece}
                \plotwitherrorareas{data/entropy-vs-SPC-cifar10-results-miniVGG-duq-combined.csv}{black}{DUQ}{mean_train_ece}{std_train_ece}
                \plotwitherrorareas{data/entropy-vs-SPC-cifar10-results-miniVGG-baseline-combined.csv}{magenta}{BL}{mean_train_ece}{std_train_ece}
                \plotwitherrorareas{data/entropy-vs-SPC-cifar10-results-miniVGG-flipout-combined.csv}{gray}{VI}{mean_train_ece}{std_train_ece}
                \plotwitherrorareas{data/maxprob-vs-SPC-cifar10-results-miniVGG-gradient-l1_norm-combined.csv}{orange}{GD}{mean_train_ece}{std_train_ece}
                \legend{}
            \end{axis}		            
        \end{tikzpicture}
        \caption{Train ECE}
    \end{subfigure}
    \begin{subfigure}{0.49\textwidth}
        \begin{tikzpicture}
            \begin{axis}[height = 0.2 \textheight, width = \textwidth, xlabel={Samples per Class}, ylabel={ECE}, ymajorgrids=false, xmajorgrids=false, grid style=dashed, legend pos = north east, legend style={font=\scriptsize}, tick label style={font=\tiny,rotate=40}, xtick=data, xmode=log, log ticks with fixed point, ymin=0.0, ymax=0.6]
                
                \plotwitherrorareas{data/entropy-vs-SPC-cifar10-results-miniVGG-deepensemble-combined.csv}{blue}{DE}{mean_ece}{std_ece}
                \plotwitherrorareas{data/entropy-vs-SPC-cifar10-results-miniVGG-dropout-combined.csv}{red}{DO}{mean_ece}{std_ece}
                \plotwitherrorareas{data/entropy-vs-SPC-cifar10-results-miniVGG-dropconnect-combined.csv}{green}{DC}{mean_ece}{std_ece}
                \plotwitherrorareas{data/entropy-vs-SPC-cifar10-results-miniVGG-duq-combined.csv}{black}{DUQ}{mean_ece}{std_ece}
                \plotwitherrorareas{data/entropy-vs-SPC-cifar10-results-miniVGG-baseline-combined.csv}{magenta}{BL}{mean_ece}{std_ece}
                \plotwitherrorareas{data/entropy-vs-SPC-cifar10-results-miniVGG-flipout-combined.csv}{gray}{VI}{mean_ece}{std_ece}
                \plotwitherrorareas{data/maxprob-vs-SPC-cifar10-results-miniVGG-gradient-l1_norm-combined.csv}{orange}{GD}{mean_ece}{std_ece}
                \legend{}
            \end{axis}		            
        \end{tikzpicture}
        \caption{Test ECE}
    \end{subfigure}
    
    \begin{subfigure}{0.49\textwidth}
        \begin{tikzpicture}
            \begin{axis}[height = 0.2 \textheight, width = \textwidth, xlabel={Samples per Class}, ylabel={OOD AUC}, ymajorgrids=false, xmajorgrids=false, grid style=dashed, legend pos = south east, legend style={font=\scriptsize}, tick label style={font=\tiny,rotate=40}, xtick=data, xmode=log, log ticks with fixed point, ymin=0.2, ymax=1.0]
                
                \plotwitherrorareas{data/entropy-vs-SPC-cifar10-results-miniVGG-deepensemble-combined.csv}{blue}{DE}{mean_ood_auc_entropy}{std_ood_auc_entropy}  
                \plotwitherrorareas{data/entropy-vs-SPC-cifar10-results-miniVGG-dropout-combined.csv}{red}{DO}{mean_ood_auc_entropy}{std_ood_auc_entropy}
                \plotwitherrorareas{data/entropy-vs-SPC-cifar10-results-miniVGG-dropconnect-combined.csv}{green}{DC}{mean_ood_auc_entropy}{std_ood_auc_entropy}
                \plotwitherrorareas{data/entropy-vs-SPC-cifar10-results-miniVGG-duq-combined.csv}{black}{DUQ}{mean_ood_auc_entropy}{std_ood_auc_entropy}
                \plotwitherrorareas{data/entropy-vs-SPC-cifar10-results-miniVGG-baseline-combined.csv}{magenta}{BL}{mean_ood_auc_entropy}{std_ood_auc_entropy}
                \plotwitherrorareas{data/entropy-vs-SPC-cifar10-results-miniVGG-flipout-combined.csv}{gray}{VI}{mean_ood_auc_entropy}{std_ood_auc_entropy}
                \legend{}
            \end{axis}		            
        \end{tikzpicture}
        \caption{Test/OOD AUC Entropy}
    \end{subfigure}
    \begin{subfigure}{0.49\textwidth}
        \begin{tikzpicture}
            \begin{axis}[height = 0.2 \textheight, width = \textwidth, xlabel={Samples per Class}, ylabel={OOD AUC}, ymajorgrids=false, xmajorgrids=false, grid style=dashed, tick label style={font=\tiny,rotate=40}, xtick=data, xmode=log, log ticks with fixed point, ymin=0.2, ymax=1.0]
                
                \plotwitherrorareas{data/entropy-vs-SPC-cifar10-results-miniVGG-deepensemble-combined.csv}{blue}{DE}{mean_ood_auc_maxprob}{std_ood_auc_maxprob}
                \plotwitherrorareas{data/entropy-vs-SPC-cifar10-results-miniVGG-dropout-combined.csv}{red}{DO}{mean_ood_auc_maxprob}{std_ood_auc_maxprob}
                \plotwitherrorareas{data/entropy-vs-SPC-cifar10-results-miniVGG-dropconnect-combined.csv}{green}{DC}{mean_ood_auc_maxprob}{std_ood_auc_maxprob}
                \plotwitherrorareas{data/entropy-vs-SPC-cifar10-results-miniVGG-duq-combined.csv}{black}{DUQ}{mean_ood_auc_maxprob}{std_ood_auc_maxprob}
                \plotwitherrorareas{data/entropy-vs-SPC-cifar10-results-miniVGG-baseline-combined.csv}{magenta}{BL}{mean_ood_auc_maxprob}{std_ood_auc_maxprob}
                \plotwitherrorareas{data/entropy-vs-SPC-cifar10-results-miniVGG-flipout-combined.csv}{gray}{VI}{mean_ood_auc_maxprob}{std_ood_auc_maxprob}
                \plotwitherrorareas{data/maxprob-vs-SPC-cifar10-results-miniVGG-gradient-l1_norm-combined.csv}{orange}{GD}{mean_ood_auc_maxprob}{std_ood_auc_maxprob}
                \legend{}
            \end{axis}		            
        \end{tikzpicture}
        \caption{Test/OOD AUC Max Prob}
    \end{subfigure}
    
    \caption{Comparison of uncertainty as size of the training set is varied on CIFAR10.}
    \label{comparison_plots_cifar10}
\end{figure*}
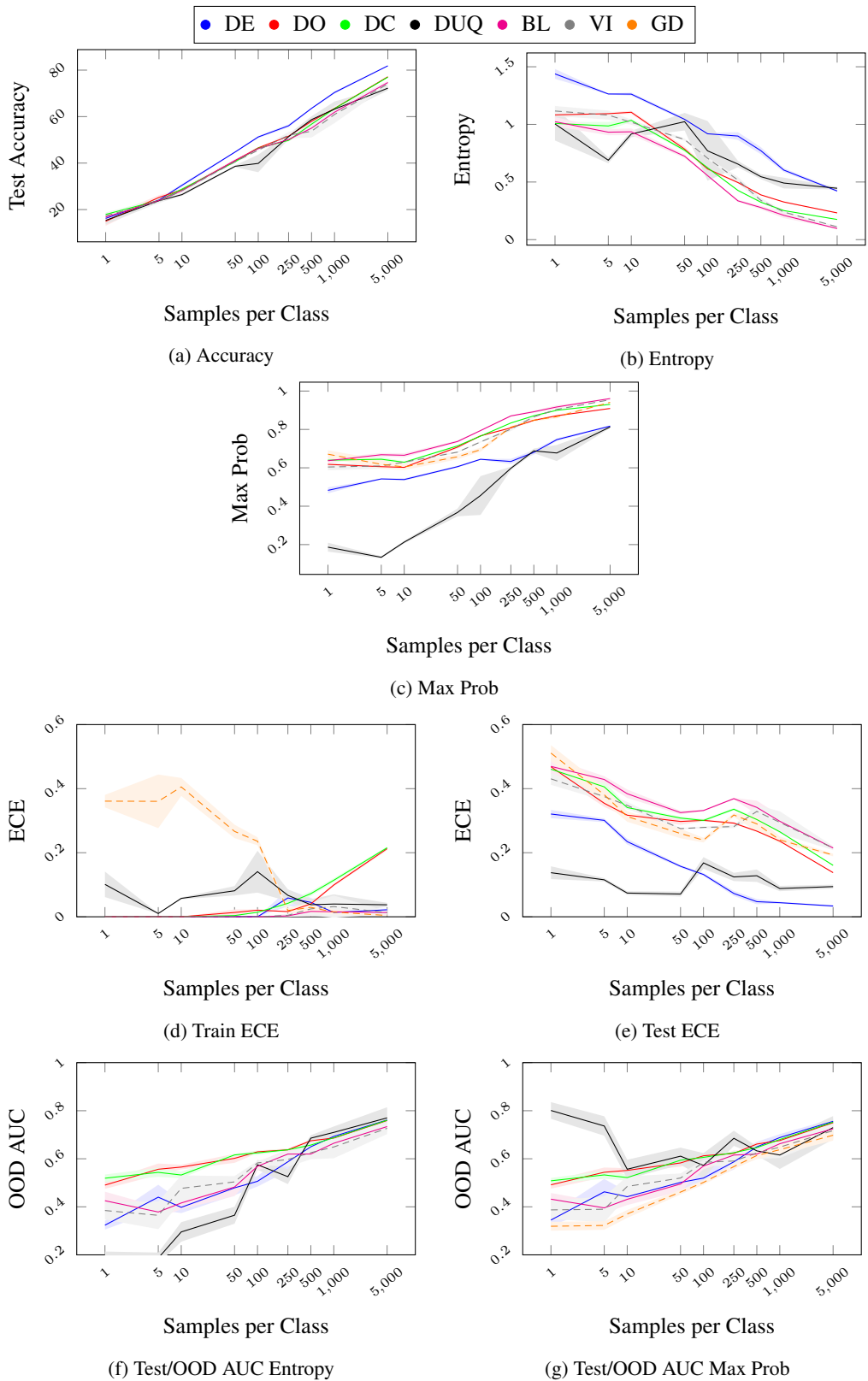

Our main results are presented in Figure \ref{comparison_plots_fmnist} for Fashion MNIST, Figure \ref{comparison_plots_cifar10} for CIFAR10, and additional out of distribution detection plots in Figures \ref{ood_comparison_fmnist} and \ref{ood_comparison_cifar10}.

In terms of accuracy, all models perform similarly, with ensembles slightly outperforming other methods in most settings, but DropConnect and Dropout are competitive in low shot settings (less than 10 sample per class).

Looking at entropy and maximum probability distributions, DUQ clearly stands out as its the least confident method in terms of maximum probability, and with higher entropy than other methods, and a peak in entropy around $\text{SPC} = 50$. All methods become more confident as the training set increases in size, which we believe indicates good estimation of epistemic uncertainty.

Calibration error shows that most methods are calibrated in the training set across all values of SPC, except for gradient uncertainty which produces uncalibrated probabilities in the low shot setting ($\text{SPC} < 250$). All methods have improved calibration on the test set as SPC increases. The best calibrated method seems to be Deep Ensembles and VI with Flipout, with DUQ producing unstable calibration error in some SPC settings. While there are no guarantees for test calibration error, overall we believe that our results show that many methods could be improved in terms of calibration with different training set sizes.

Out of distribution performance is more varied. In the Test/OOD experiment, all uncertainty methods improve their performance as SPC increases, and this is more evident with entropy than maximum probability. In the Train/Test experiment, all methods start with high AUC when SPC is low, and constantly decrease as SPC increases, signaling that uncertainty cannot be used to discriminate between train and test sets. We think this makes sense and validates the changes in uncertainty as the training set size is varied. With a small training set, it is easier to discriminate the test set since the classifier has not learned good class concepts and it is uncertain enough on the test set.

Finally, the Train/OOD experiment shows that with low SPC it is very easy to discriminate the OOD dataset, since it is very different than the training set knowledge, but this AUC performance decreases with most methods when SPC increases. Only DUQ and Gradient perform in the opposite way. Ensembles in this scenario has almost constant performance. The main takeaways from our experimental results are:

\begin{itemize}
    \item All methods except gradient, across all training set sizes, are well calibrated in the training set, but miscalibrated on the test set, with improving calibration as the training set size increases (SPC $\uparrow$).
    \item DUQ is considerably less confident when SPC is low, which indicates that it correctly gauges its own uncertainty, but these results are mixed when looking at test calibration error when $\text{SPC} < 50$.
    \item Gradient-based methods seems to estimate a very different kind of uncertainty than other methods, with poor Test/OOD AUC performance, and the worse calibration error both in train and test sets. More research is needed in order to understand how these methods work and the effect of their aggregation metric (see Figure \ref{grad_unc_comparison}).
    \item Ensembles are competitive in terms of accuracy and calibration error, but do not perform as well in some out of distribution detection scenarios (Test/OOD).
    \item It is not clear if maximum probability or entropy is the best for out of distribution detection, performance varies considerably between the two. It should be chosen carefully \cite{hendrycks2017baseline}.
    \item There is no method that clearly outperforms all others across varied SPC values. Some methods work very well for calibration, but are outperformed in different out of distribution detection settings.
    \item Selection of uncertainty methods should be done carefully, considering the training set size for a given task.
\end{itemize}

\section{Conclusions and Future Work}

In this paper we evaluated the performance of various uncertainty quantification methods as the training set size is varied. We perform a comprehensive evaluation in terms of metrics for uncertainty, including calibration error and out of distribution performance as measured by AUC in several settings. Our overall results show that uncertainty quantification performcance has a strong dependency on the training set size, which can be related to model uncertainty, but this relationship is not always intuitive or predictable. We make clear takeaways for the community to learn from our results.

We expect that our results can guide future research into these methods, in particular for DUQ and gradient-based methods. DUQ provides excellent epistemic uncertainty quantification, while we believe that gradient-based methods do not estimate epistemic uncertainty correctly.

We believe our results can guide practitioners into selecting proper methods to estimate uncertainty of machine learning models.

\FloatBarrier
\clearpage

\bibliographystyle{plain}
\bibliography{biblio}

\begin{thebibliography}{10}

\bibitem{blundell2015weight}
Charles Blundell, Julien Cornebise, Koray Kavukcuoglu, and Daan Wierstra.
\newblock Weight uncertainty in neural network.
\newblock In {\em International Conference on Machine Learning}, pages
  1613--1622. PMLR, 2015.

\bibitem{der2009aleatory}
Armen Der~Kiureghian and Ove Ditlevsen.
\newblock Aleatory or epistemic? does it matter?
\newblock {\em Structural safety}, 31(2):105--112, 2009.

\bibitem{feng2021review}
Di~Feng, Ali Harakeh, Steven~L Waslander, and Klaus Dietmayer.
\newblock A review and comparative study on probabilistic object detection in
  autonomous driving.
\newblock {\em IEEE Transactions on Intelligent Transportation Systems}, 2021.

\bibitem{gal2016dropout}
Yarin Gal and Zoubin Ghahramani.
\newblock Dropout as a bayesian approximation: Representing model uncertainty
  in deep learning.
\newblock In {\em international conference on machine learning}, pages
  1050--1059. PMLR, 2016.

\bibitem{gawlikowski2021survey}
Jakob Gawlikowski, Cedrique Rovile~Njieutcheu Tassi, Mohsin Ali, Jongseok Lee,
  Matthias Humt, Jianxiang Feng, Anna Kruspe, Rudolph Triebel, Peter Jung,
  Ribana Roscher, et~al.
\newblock A survey of uncertainty in deep neural networks.
\newblock {\em arXiv preprint arXiv:2107.03342}, 2021.

\bibitem{guo2017calibration}
Chuan Guo, Geoff Pleiss, Yu~Sun, and Kilian~Q Weinberger.
\newblock On calibration of modern neural networks.
\newblock In {\em International Conference on Machine Learning}, pages
  1321--1330. PMLR, 2017.

\bibitem{hendrycks2017baseline}
Dan Hendrycks and Kevin Gimpel.
\newblock A baseline for detecting misclassified and out-of-distribution
  examples in neural networks.
\newblock In {\em International Conference on Learning Representations}, 2017.

\bibitem{hullermeier2021aleatoric}
Eyke H{\"u}llermeier and Willem Waegeman.
\newblock Aleatoric and epistemic uncertainty in machine learning: An
  introduction to concepts and methods.
\newblock {\em Machine Learning}, 110(3):457--506, 2021.

\bibitem{kendall2017uncertainties}
Alex Kendall and Yarin Gal.
\newblock What uncertainties do we need in bayesian deep learning for computer
  vision?
\newblock {\em arXiv preprint arXiv:1703.04977}, 2017.

\bibitem{lakshminarayanan2017simple}
Balaji Lakshminarayanan, Alexander Pritzel, and Charles Blundell.
\newblock Simple and scalable predictive uncertainty estimation using deep
  ensembles.
\newblock In {\em Advances in Neural Information Processing Systems}, pages
  6402--6413, 2017.

\bibitem{mobiny2021dropconnect}
Aryan Mobiny, Pengyu Yuan, Supratik~K Moulik, Naveen Garg, Carol~C Wu, and Hien
  Van~Nguyen.
\newblock Dropconnect is effective in modeling uncertainty of bayesian deep
  networks.
\newblock {\em Scientific reports}, 11(1):1--14, 2021.

\bibitem{oberdiek2018classification}
Philipp Oberdiek, Matthias Rottmann, and Hanno Gottschalk.
\newblock Classification uncertainty of deep neural networks based on gradient
  information.
\newblock In {\em IAPR Workshop on Artificial Neural Networks in Pattern
  Recognition}, pages 113--125. Springer, 2018.

\bibitem{pedregosa2011scikit}
Fabian Pedregosa, Ga{\"e}l Varoquaux, Alexandre Gramfort, Vincent Michel,
  Bertrand Thirion, Olivier Grisel, Mathieu Blondel, Peter Prettenhofer, Ron
  Weiss, Vincent Dubourg, et~al.
\newblock Scikit-learn: Machine learning in python.
\newblock {\em the Journal of machine Learning research}, 12:2825--2830, 2011.

\bibitem{sunderhauf2018limits}
Niko S{\"u}nderhauf, Oliver Brock, Walter Scheirer, Raia Hadsell, Dieter Fox,
  J{\"u}rgen Leitner, Ben Upcroft, Pieter Abbeel, Wolfram Burgard, Michael
  Milford, et~al.
\newblock The limits and potentials of deep learning for robotics.
\newblock {\em The International Journal of Robotics Research},
  37(4-5):405--420, 2018.

\bibitem{van2020uncertainty}
Joost Van~Amersfoort, Lewis Smith, Yee~Whye Teh, and Yarin Gal.
\newblock Uncertainty estimation using a single deep deterministic neural
  network.
\newblock In {\em International Conference on Machine Learning}, pages
  9690--9700. PMLR, 2020.

\bibitem{wen2018flipout}
Yeming Wen, Paul Vicol, Jimmy Ba, Dustin Tran, and Roger Grosse.
\newblock Flipout: Efficient pseudo-independent weight perturbations on
  mini-batches.
\newblock {\em arXiv preprint arXiv:1803.04386}, 2018.

\bibitem{wilson2020case}
Andrew~Gordon Wilson.
\newblock The case for bayesian deep learning.
\newblock {\em arXiv preprint arXiv:2001.10995}, 2020.

\end{thebibliography}

\clearpage
\appendix

\section{Classification Neural Network Architecture and Training Setup}

All models over all datasets are trained using the same convolutional architecture. The purpose is to get comparable results, not to obtain performance that mimics the state of the art.

The convolutional architecture uses convolution with 64 $3 \times 3$ filters, followed by $2 \times 2$ Max-Pooling, then 128 $3 \times 3$ filters with $2 \times 2$ Max-Pooling, and finally 128 $3 \times 3$ filters with $2 \times 2$ Max-Pooling. The network is complete with two fully connected layers, one with 256 units, and the output layer with $C$ units equal to the number of classes, and a softmax activation. All layers except the output use a ReLU activation, and we insert Batch Normalization layers between Convolutional and Max-Pooling layers.

This network without any uncertainty quantification method applied is denoted as Baseline (BL) in our experiments.

\section{Details on Uncertainty Methods for Classification}

In this section we briefly describe the uncertainty method we used, their hyper-parameters, and any modifications that we made to their training procedures.

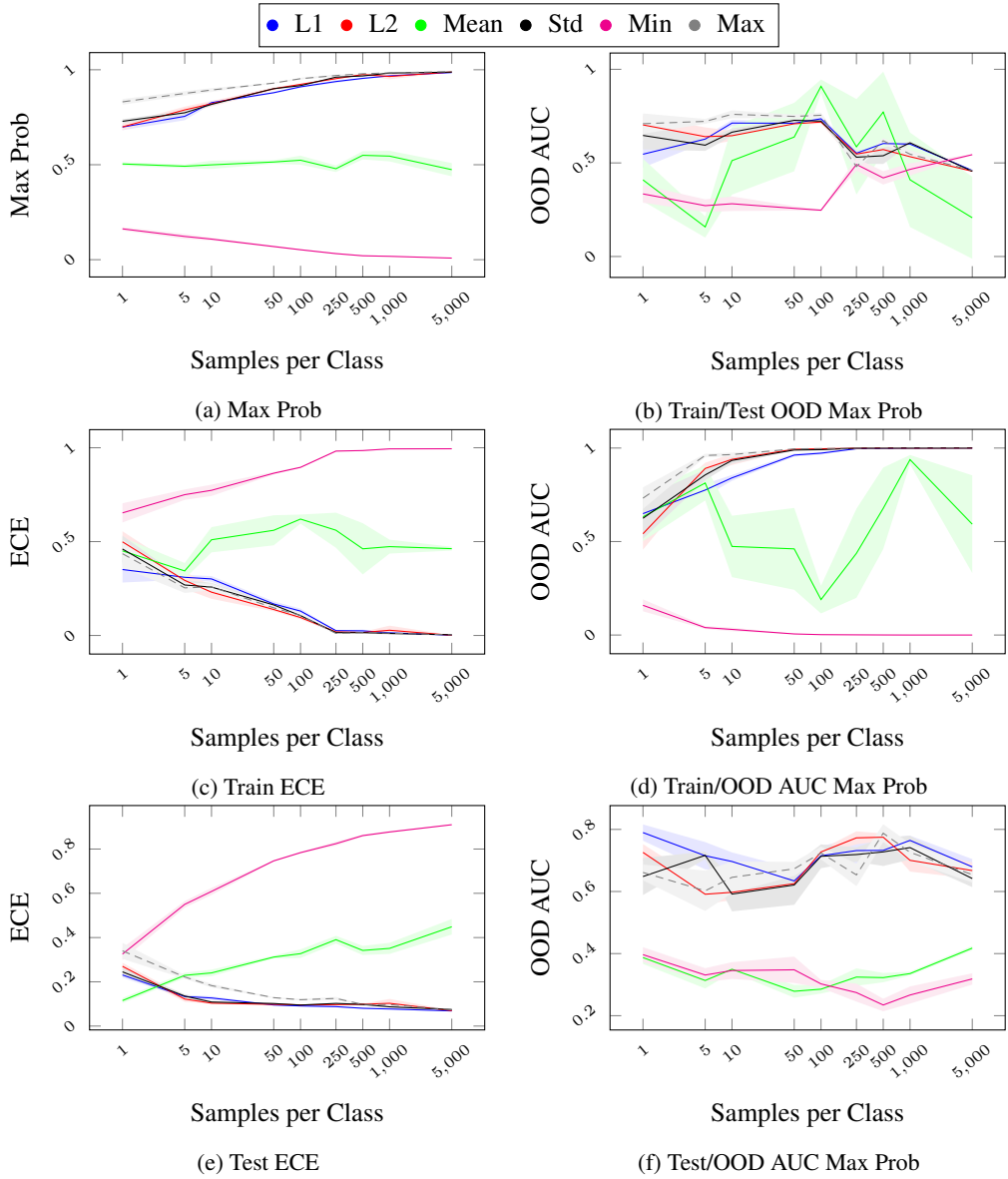
\begin{figure}[th]
    \centering
    \begin{tikzpicture}
        \begin{customlegend}[legend columns = 7,legend style = {column sep=1ex}, legend cell align = left,
            legend entries={L1, L2, Mean, Std, Min, Max}]
            \addlegendimage{mark=none,blue, only marks}
            \addlegendimage{mark=none,red, only marks}
            \addlegendimage{mark=none,green, only marks}
            \addlegendimage{mark=none,black, only marks}
            \addlegendimage{mark=none,magenta, only marks}
            \addlegendimage{mark=none,gray, only marks}
            \addlegendimage{mark=none,orange, only marks}
        \end{customlegend}
    \end{tikzpicture}

    \begin{subfigure}{0.49\textwidth}
        \begin{tikzpicture}
            \begin{axis}[height = 0.2 \textheight, width = \textwidth, xlabel={Samples per Class}, ylabel={Max Prob}, ymajorgrids=false, xmajorgrids=false, grid style=dashed, legend pos = south east, legend style={font=\scriptsize}, tick label style={font=\tiny,rotate=40}, xtick=data, xmode=log, log ticks with fixed point]               
                \plotwitherrorareas{data/maxprob-vs-SPC-fashion_mnist-results-miniVGG-gradient-l1_norm-combined.csv}{blue}{DE}{mean_mean_maxprob}{std_mean_maxprob}
                \plotwitherrorareas{data/maxprob-vs-SPC-fashion_mnist-results-miniVGG-gradient-l2_norm-combined.csv}{red}{DO}{mean_mean_maxprob}{std_mean_maxprob}
                \plotwitherrorareas{data/maxprob-vs-SPC-fashion_mnist-results-miniVGG-gradient-mean-combined.csv}{green}{DC}{mean_mean_maxprob}{std_mean_maxprob}
                \plotwitherrorareas{data/maxprob-vs-SPC-fashion_mnist-results-miniVGG-gradient-std-combined.csv}{black}{DUQ}{mean_mean_maxprob}{std_mean_maxprob}
                \plotwitherrorareas{data/maxprob-vs-SPC-fashion_mnist-results-miniVGG-gradient-min-combined.csv}{magenta}{BL}{mean_mean_maxprob}{std_mean_maxprob}
                \plotwitherrorareas{data/maxprob-vs-SPC-fashion_mnist-results-miniVGG-gradient-max-combined.csv}{gray}{VI}{mean_mean_maxprob}{std_mean_maxprob}
                \legend{}
            \end{axis}		            
jja        \end{tikzpicture}
        \caption{Max Prob}
    \end{subfigure}
    \begin{subfigure}{0.49\textwidth}
        \begin{tikzpicture}
            \begin{axis}[height = 0.2 \textheight, width = \textwidth, xlabel={Samples per Class}, ylabel={OOD AUC}, ymajorgrids=false, xmajorgrids=false, grid style=dashed, tick label style={font=\tiny,rotate=40}, xtick=data, xmode=log, log ticks with fixed point]                
                \plotwitherrorareas{data/maxprob-vs-SPC-fashion_mnist-results-miniVGG-gradient-l1_norm-combined.csv}{blue}{DE}{mean_tr_test_auc_maxprob}{std_tr_test_auc_maxprob}
                \plotwitherrorareas{data/maxprob-vs-SPC-fashion_mnist-results-miniVGG-gradient-l2_norm-combined.csv}{red}{DO}{mean_tr_test_auc_maxprob}{std_tr_test_auc_maxprob}
                \plotwitherrorareas{data/maxprob-vs-SPC-fashion_mnist-results-miniVGG-gradient-mean-combined.csv}{green}{DC}{mean_tr_test_auc_maxprob}{std_tr_test_auc_maxprob}
                \plotwitherrorareas{data/maxprob-vs-SPC-fashion_mnist-results-miniVGG-gradient-std-combined.csv}{black}{DUQ}{mean_tr_test_auc_maxprob}{std_tr_test_auc_maxprob}
                \plotwitherrorareas{data/maxprob-vs-SPC-fashion_mnist-results-miniVGG-gradient-min-combined.csv}{magenta}{BL}{mean_tr_test_auc_maxprob}{std_tr_test_auc_maxprob}
                \plotwitherrorareas{data/maxprob-vs-SPC-fashion_mnist-results-miniVGG-gradient-max-combined.csv}{gray}{VI}{mean_tr_test_auc_maxprob}{std_tr_test_auc_maxprob}
                \legend{}
            \end{axis}
        \end{tikzpicture}
        \caption{Train/Test OOD Max Prob}
    \end{subfigure}
    
    \begin{subfigure}{0.49\textwidth}
        \begin{tikzpicture}
            \begin{axis}[height = 0.2 \textheight, width = \textwidth, xlabel={Samples per Class}, ylabel={ECE}, ymajorgrids=false, xmajorgrids=false, grid style=dashed, legend pos = north east, legend style={font=\scriptsize}, tick label style={font=\tiny,rotate=40}, xtick=data, xmode=log, log ticks with fixed point]                
                \plotwitherrorareas{data/maxprob-vs-SPC-fashion_mnist-results-miniVGG-gradient-l1_norm-combined.csv}{blue}{DE}{mean_train_ece}{std_train_ece}
                \plotwitherrorareas{data/maxprob-vs-SPC-fashion_mnist-results-miniVGG-gradient-l2_norm-combined.csv}{red}{DO}{mean_train_ece}{std_train_ece}
                \plotwitherrorareas{data/maxprob-vs-SPC-fashion_mnist-results-miniVGG-gradient-mean-combined.csv}{green}{DC}{mean_train_ece}{std_train_ece}
                \plotwitherrorareas{data/maxprob-vs-SPC-fashion_mnist-results-miniVGG-gradient-std-combined.csv}{black}{DUQ}{mean_train_ece}{std_train_ece}
                \plotwitherrorareas{data/maxprob-vs-SPC-fashion_mnist-results-miniVGG-gradient-min-combined.csv}{magenta}{BL}{mean_train_ece}{std_train_ece}
                \plotwitherrorareas{data/maxprob-vs-SPC-fashion_mnist-results-miniVGG-gradient-max-combined.csv}{gray}{VI}{mean_train_ece}{std_train_ece}
                \legend{}
            \end{axis}		            
        \end{tikzpicture}
        \caption{Train ECE}
    \end{subfigure}
    \begin{subfigure}{0.49\textwidth}
        \begin{tikzpicture}
            \begin{axis}[height = 0.2 \textheight, width = \textwidth, xlabel={Samples per Class}, ylabel={OOD AUC}, ymajorgrids=false, xmajorgrids=false, grid style=dashed, legend pos = south east, legend style={font=\scriptsize}, tick label style={font=\tiny,rotate=40}, xtick=data, xmode=log, log ticks with fixed point]
                
                \plotwitherrorareas{data/maxprob-vs-SPC-fashion_mnist-results-miniVGG-gradient-l1_norm-combined.csv}{blue}{DE}{mean_tr_ood_auc_maxprob}{std_tr_ood_auc_maxprob}
                \plotwitherrorareas{data/maxprob-vs-SPC-fashion_mnist-results-miniVGG-gradient-l2_norm-combined.csv}{red}{DO}{mean_tr_ood_auc_maxprob}{std_tr_ood_auc_maxprob}
                \plotwitherrorareas{data/maxprob-vs-SPC-fashion_mnist-results-miniVGG-gradient-mean-combined.csv}{green}{DC}{mean_tr_ood_auc_maxprob}{std_tr_ood_auc_maxprob}
                \plotwitherrorareas{data/maxprob-vs-SPC-fashion_mnist-results-miniVGG-gradient-std-combined.csv}{black}{DUQ}{mean_tr_ood_auc_maxprob}{std_tr_ood_auc_maxprob}
                \plotwitherrorareas{data/maxprob-vs-SPC-fashion_mnist-results-miniVGG-gradient-min-combined.csv}{magenta}{BL}{mean_tr_ood_auc_maxprob}{std_tr_ood_auc_maxprob}
                \plotwitherrorareas{data/maxprob-vs-SPC-fashion_mnist-results-miniVGG-gradient-max-combined.csv}{gray}{VI}{mean_tr_ood_auc_maxprob}{std_tr_ood_auc_maxprob}                
                \legend{}
            \end{axis}		            
        \end{tikzpicture}
        \caption{Train/OOD AUC Max Prob}
    \end{subfigure}
    
    \begin{subfigure}{0.49\textwidth}
        \begin{tikzpicture}
            \begin{axis}[height = 0.2 \textheight, width = \textwidth, xlabel={Samples per Class}, ylabel={ECE}, ymajorgrids=false, xmajorgrids=false, grid style=dashed, legend pos = north east, legend style={font=\scriptsize}, tick label style={font=\tiny,rotate=40}, xtick=data, xmode=log, log ticks with fixed point]                
                \plotwitherrorareas{data/maxprob-vs-SPC-fashion_mnist-results-miniVGG-gradient-l1_norm-combined.csv}{blue}{DE}{mean_ece}{std_ece}
                \plotwitherrorareas{data/maxprob-vs-SPC-fashion_mnist-results-miniVGG-gradient-l2_norm-combined.csv}{red}{DO}{mean_ece}{std_ece}
                \plotwitherrorareas{data/maxprob-vs-SPC-fashion_mnist-results-miniVGG-gradient-mean-combined.csv}{green}{DC}{mean_ece}{std_ece}
                \plotwitherrorareas{data/maxprob-vs-SPC-fashion_mnist-results-miniVGG-gradient-std-combined.csv}{black}{DUQ}{mean_ece}{std_ece}
                \plotwitherrorareas{data/maxprob-vs-SPC-fashion_mnist-results-miniVGG-gradient-min-combined.csv}{magenta}{BL}{mean_ece}{std_ece}
                \plotwitherrorareas{data/maxprob-vs-SPC-fashion_mnist-results-miniVGG-gradient-max-combined.csv}{gray}{VI}{mean_ece}{std_ece}
                \legend{}
            \end{axis}		            
        \end{tikzpicture}
        \caption{Test ECE}
    \end{subfigure}    
    \begin{subfigure}{0.49\textwidth}
        \begin{tikzpicture}
            \begin{axis}[height = 0.2 \textheight, width = \textwidth, xlabel={Samples per Class}, ylabel={OOD AUC}, ymajorgrids=false, xmajorgrids=false, grid style=dashed, tick label style={font=\tiny,rotate=40}, xtick=data, xmode=log, log ticks with fixed point]
                
                \plotwitherrorareas{data/maxprob-vs-SPC-fashion_mnist-results-miniVGG-gradient-l1_norm-combined.csv}{blue}{DE}{mean_ood_auc_maxprob}{std_ood_auc_maxprob}
                \plotwitherrorareas{data/maxprob-vs-SPC-fashion_mnist-results-miniVGG-gradient-l2_norm-combined.csv}{red}{DO}{mean_ood_auc_maxprob}{std_ood_auc_maxprob}
                \plotwitherrorareas{data/maxprob-vs-SPC-fashion_mnist-results-miniVGG-gradient-mean-combined.csv}{green}{DC}{mean_ood_auc_maxprob}{std_ood_auc_maxprob}
                \plotwitherrorareas{data/maxprob-vs-SPC-fashion_mnist-results-miniVGG-gradient-std-combined.csv}{black}{DUQ}{mean_ood_auc_maxprob}{std_ood_auc_maxprob}
                \plotwitherrorareas{data/maxprob-vs-SPC-fashion_mnist-results-miniVGG-gradient-min-combined.csv}{magenta}{BL}{mean_ood_auc_maxprob}{std_ood_auc_maxprob}
                \plotwitherrorareas{data/maxprob-vs-SPC-fashion_mnist-results-miniVGG-gradient-max-combined.csv}{gray}{VI}{mean_ood_auc_maxprob}{std_ood_auc_maxprob}                
                \legend{}
            \end{axis}		            
        \end{tikzpicture}
        \caption{Test/OOD AUC Max Prob}
    \end{subfigure}
    \caption{Comparison of Gradient uncertainty on Fashion MNIST across training set sizes, with different aggregation metrics.}
    \label{grad_unc_comparison}
\end{figure}

\begin{description}
    \item[MC-Dropout (DO)] Dropout sets random activations in a layer to zero, and it is intended as a regularizer that is only applied during training. MC-Dropout \cite{gal2016dropout} enables this activation drop during test/inference time, and the model becomes stochastic, where each forward pass produces one sample from the Bayesian posterior distribution \cite{gal2016dropout}. We use one dropout layer before the last layer, with a drop probability $p = 0.25$. For evaluation we take $M = 50$ forward passes and take the mean over samples as the output prediction.
    \item[MC-DropConnect (DC)] DropConnect is very similar to Dropout, with the difference being that DropConnect randomly sets weights to zero instead of activations, with the same regularizing effect. MC-DropConnect enables DropConnect at inference time, and it has also been shown to produce samples from the Bayesian posterior distribution \cite{mobiny2021dropconnect}. We use a single DropConnect layer at the network output (replacing the standard Dense layer) with drop probability $p = 0.25$. For evaluation we take $M = 50$ forward passes and take the mean over samples as the output prediction.
    \item[Deep Ensembles (DE)] Ensembling is a standard method in Machine Learning, where outputs of several models are combined, which usually produces a better model. It has been shown \cite{lakshminarayanan2017simple} that ensembles also have excellent uncertainty quantification properties. We use an ensemble of neural networks with the same architecture and $N = 5$ ensemble members.
    \item[Direct Uncertainty Quantification (DUQ)] This method \cite{van2020uncertainty} replaces the standard softmax classifier with a radial basis function (RBF) classifier, where the output layer learns a weight matrix and a centroid for each class, using the minimum distance to a centroid to decide which class to output, and the centroid distance as an uncertainty measure. Centroids are updated using a running mean on input feature space, but we found that this method is unstable and does not converge in a low-show setting, so we learn the centroids using gradient descent. 
    \item[VI with Flipout (VI)] Variational inference is a popular method which models weights as approximate distributions, usually selecting a Gaussian distribution \cite{blundell2015weight}, where the components of the kernel and bias matrices are Gaussian distributions.
    This transforms the model into a stochastic one. Flipout \cite{wen2018flipout} is used as an additional formulation on top of a stochastic perturbation model that reduces variance, greatly improving learning stability and performance. To effectively learn across different training set sizes, we disable the use of a prior, and only the weights are modeled as a Gaussian distribution, the bias being a fixed learnable scalar value. For evaluation we take $M = 50$ forward passes and take the mean over samples as the output prediction. Our network architecture uses Flipout VI only in the output layer.
    \item[Gradient Uncertainty (GD)] This method \cite{oberdiek2018classification} computes the gradient of the loss with respect to trainable parameters, using a virtual label that is the one-hot encoded version of the predicted label, and passes the gradient vector through an aggregation function that produces a scalar, which can be used as an uncertainty measure. This can only be done in a classification setting.
    We normalize the aggregated gradient $g$ to the $[0, 1]$ range using min-max normalization and transform it to a pseudo probability as $p = 1- g$, which can be used to evaluate calibration error and maximum probability metrics. We found that the aggregation metric has a large impact on performance across training set sizes, as shown in Figure \ref{grad_unc_comparison}, and overall we use the L1 metric as it performs the best on most metrics (as shown in Figure \ref{grad_unc_comparison}). Note that this method produces a single confidence value for a prediction.
\end{description}

\section{Code Implementation}

Source code implementation is available at \url{https://github.com/mvaldenegro/paper-quality-epistemic-uncertainty-bayes}.

This implementation uses Keras 2.2.4, TensorFlow 1.14, and Keras-Uncertainty (available at \url{https://github.com/mvaldenegro/keras-uncertainty}).

Models were trained on a single RTX 2070 GPU. Each model instance takes from 1 to 15 minutes to train (depending on training set size), with all experiments taking less than 24 hours of GPU time.

\section{Additional Regression Toy Example}
\label{sec:toy_reg}

\begin{figure}[!htb]
    \centering
    \includegraphics[width=0.99\textwidth]{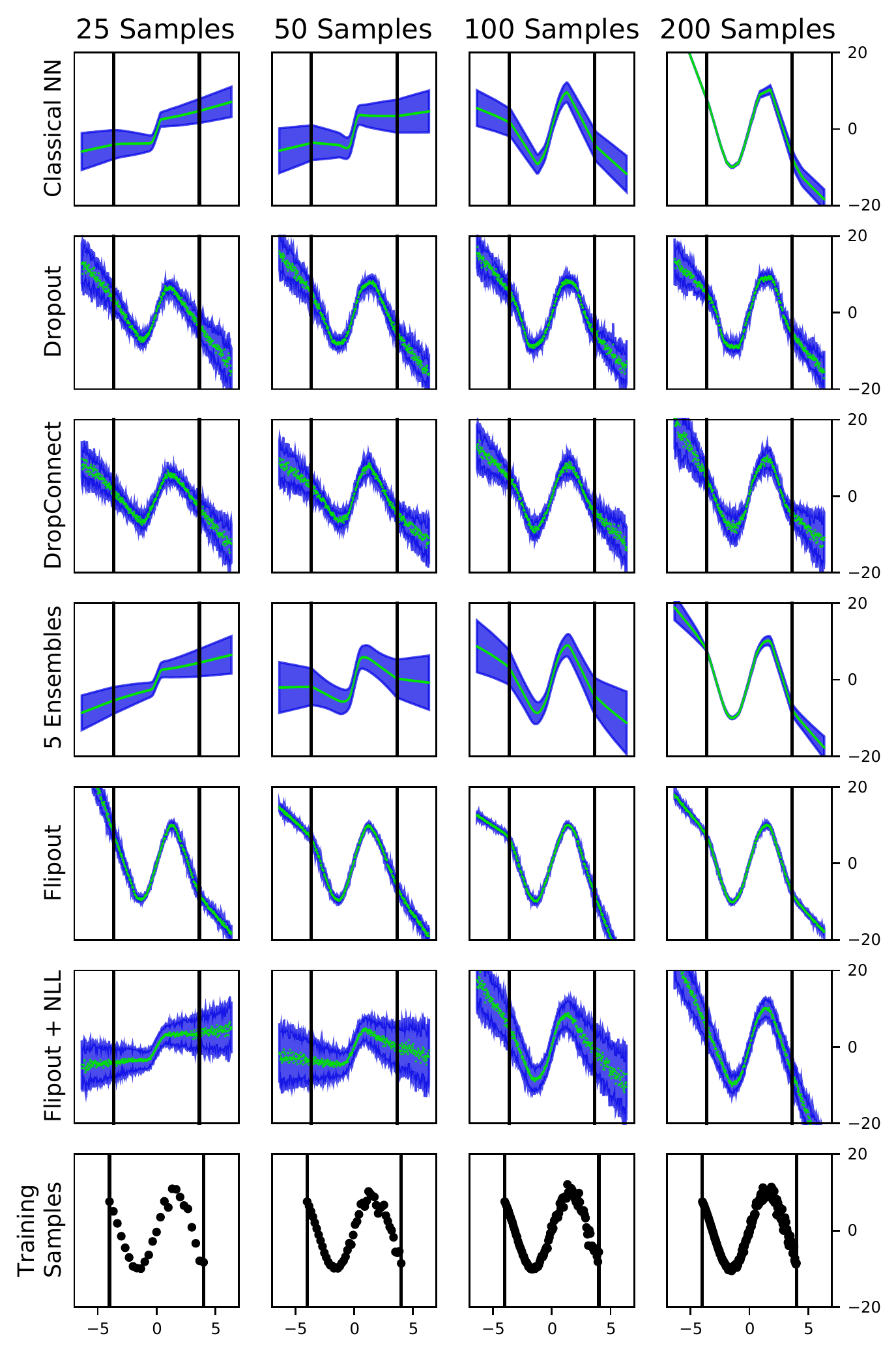}
    \caption{Comparison of uncertainty in the toy regression example, as number of training samples is varied. The last row shows the samples used for training. The two black lines indicate the limits of the training set, while the out of distribution test set ranges at $[-7, -4] \cup [4, 7]$. Green represents the mean, while the blue shaded areas are one standard deviation uncertainty. These plots clearly show that uncertainty degrades with small training sets.}
    \label{toy_reg_example}
\end{figure}

In this section we present a small regression toy example, corresponding to sampling the function $f(x) = \sin(x) + \epsilon$, where $\epsilon \sim \mathcal{N}(0, \sigma(x))$ and $\sigma(x) = 0.15(1 + e^{-x})^{-1}$ where $x \in [-4, 4]$. The interval $[-4, 4]$ is sampled at $S \in [25, 50, 100, 200]$ equally spaced samples, and various models are trained using a mean squared error loss. Evaluation happens on a fixed sized set at ranges $[-7, 7]$. This result is presented in Figure \ref{toy_reg_example}.

For the regression setting, we use a Negative Log-Likelihood (NLL) loss to capture epistemic uncertainty \cite{lakshminarayanan2017simple} \cite{kendall2017uncertainties}, which is formulated below:

\begin{equation}
    L(\mathbf{x}, y) = 0.5 N^{-1}\sum_i \left(\log \sigma^2(\mathbf{x}_i) + \frac{(\mu(\mathbf{x}_i) - y_i)^2}{\sigma^2(\mathbf{x}_i)}\right)
\end{equation}

This loss requires that the model contains two output heads, one for the mean $\mu_i(\mathbf{x})$ and another for the variance $\sigma^2(\mathbf{x})$. The variance head uses a softplus activation function in order to always predict positive soft variances.

With this loss, the output variance $\sigma^2(x)$ can be interpreted as an estimate of the aleatoric uncertainty in the data. Epistemic uncertainty is computed through a specific uncertainty quantification method.

The network architecture is two fully connected layers with 32 units each, and a ReLU activation, and two output heads (mean and variance) with a single neuron each, and a linear activation for the mean head. No Batch Normalization layers are used in this example. Models that do not predict variance use only a single output head.

Figure \ref{toy_reg_example} shows that Flipout without NLL (using mean squared error instead) cannot estimate aleatoric uncertainty, while using the NLL loss the model can estimate aleatoric uncertainty correctly. Classical NN, Ensembles, and Flipout + NLL do not correctly fit the training set with a low number of samples (less than 100 samples), while Flipout and MC Dropout and MC DropConnect do estimate the training function more closely. For Flipout + NLL and Ensembles, their large uncertainty indicates that it is an incorrect fit and shows that their uncertainty is useful.

\clearpage
\FloatBarrier
\section{Out of Distribution Detection Performance}

\begin{figure*}[!ht]
    \begin{subfigure}{0.49\textwidth}
        \begin{tikzpicture}
            \begin{axis}[height = 0.18 \textheight, width = \textwidth, xlabel={Samples per Class}, ylabel={OOD AUC}, ymajorgrids=false, xmajorgrids=false, grid style=dashed, legend pos = south east, legend style={font=\scriptsize}, tick label style={font=\tiny,rotate=40}, xtick=data, xmode=log, log ticks with fixed point, ymin=0.2, ymax=1.0]
                
                \plotwitherrorareas{data/entropy-vs-SPC-fashion_mnist-results-miniVGG-deepensemble-combined.csv}{blue}{DE}{mean_tr_test_auc_entropy}{std_tr_test_auc_entropy}  
                \plotwitherrorareas{data/entropy-vs-SPC-fashion_mnist-results-miniVGG-dropout-combined.csv}{red}{DO}{mean_tr_test_auc_entropy}{std_tr_test_auc_entropy}
                \plotwitherrorareas{data/entropy-vs-SPC-fashion_mnist-results-miniVGG-dropconnect-combined.csv}{green}{DC}{mean_tr_test_auc_entropy}{std_tr_test_auc_entropy}
                \plotwitherrorareas{data/entropy-vs-SPC-fashion_mnist-results-miniVGG-duq-combined.csv}{black}{DUQ}{mean_tr_test_auc_entropy}{std_tr_test_auc_entropy}
                \plotwitherrorareas{data/entropy-vs-SPC-fashion_mnist-results-miniVGG-baseline-combined.csv}{magenta}{BL}{mean_tr_test_auc_entropy}{std_tr_test_auc_entropy}
                \plotwitherrorareas{data/entropy-vs-SPC-fashion_mnist-results-miniVGG-flipout-combined.csv}{gray}{VI}{mean_tr_test_auc_entropy}{std_tr_test_auc_entropy}
                \legend{}
            \end{axis}		            
        \end{tikzpicture}
        \caption{Train/Test AUC Entropy}
    \end{subfigure}
    \begin{subfigure}{0.49\textwidth}
        \begin{tikzpicture}
            \begin{axis}[height = 0.18 \textheight, width = \textwidth, xlabel={Samples per Class}, ylabel={OOD AUC}, ymajorgrids=false, xmajorgrids=false, grid style=dashed, tick label style={font=\tiny,rotate=40}, xtick=data, xmode=log, log ticks with fixed point, ymin=0.2, ymax=1.0]
                
                \plotwitherrorareas{data/entropy-vs-SPC-fashion_mnist-results-miniVGG-deepensemble-combined.csv}{blue}{DE}{mean_tr_test_auc_maxprob}{std_tr_test_auc_maxprob}
                \plotwitherrorareas{data/entropy-vs-SPC-fashion_mnist-results-miniVGG-dropout-combined.csv}{red}{DO}{mean_tr_test_auc_maxprob}{std_tr_test_auc_maxprob}
                \plotwitherrorareas{data/entropy-vs-SPC-fashion_mnist-results-miniVGG-dropconnect-combined.csv}{green}{DC}{mean_tr_test_auc_maxprob}{std_tr_test_auc_maxprob}
                \plotwitherrorareas{data/entropy-vs-SPC-fashion_mnist-results-miniVGG-duq-combined.csv}{black}{DUQ}{mean_tr_test_auc_maxprob}{std_tr_test_auc_maxprob}
                \plotwitherrorareas{data/entropy-vs-SPC-fashion_mnist-results-miniVGG-baseline-combined.csv}{magenta}{BL}{mean_tr_test_auc_maxprob}{std_tr_test_auc_maxprob}
                \plotwitherrorareas{data/entropy-vs-SPC-fashion_mnist-results-miniVGG-flipout-combined.csv}{gray}{VI}{mean_tr_test_auc_maxprob}{std_tr_test_auc_maxprob}
                \plotwitherrorareas{data/maxprob-vs-SPC-fashion_mnist-results-miniVGG-gradient-l1_norm-combined.csv}{orange}{GD}{mean_tr_test_auc_maxprob}{std_tr_test_auc_maxprob}
                \legend{}
            \end{axis}		            
        \end{tikzpicture}
        \caption{Train/Test AUC Max Prob}
    \end{subfigure}
    
    \begin{subfigure}{0.49\textwidth}
        \begin{tikzpicture}
            \begin{axis}[height = 0.18 \textheight, width = \textwidth, xlabel={Samples per Class}, ylabel={OOD AUC}, ymajorgrids=false, xmajorgrids=false, grid style=dashed, legend pos = south east, legend style={font=\scriptsize}, tick label style={font=\tiny,rotate=40}, xtick=data, xmode=log, log ticks with fixed point, ymin=0.2, ymax=1.0]
                
                \plotwitherrorareas{data/entropy-vs-SPC-fashion_mnist-results-miniVGG-deepensemble-combined.csv}{blue}{DE}{mean_tr_ood_auc_entropy}{std_tr_ood_auc_entropy}  
                \plotwitherrorareas{data/entropy-vs-SPC-fashion_mnist-results-miniVGG-dropout-combined.csv}{red}{DO}{mean_tr_ood_auc_entropy}{std_tr_ood_auc_entropy}
                \plotwitherrorareas{data/entropy-vs-SPC-fashion_mnist-results-miniVGG-dropconnect-combined.csv}{green}{DC}{mean_tr_ood_auc_entropy}{std_tr_ood_auc_entropy}
                \plotwitherrorareas{data/entropy-vs-SPC-fashion_mnist-results-miniVGG-duq-combined.csv}{black}{DUQ}{mean_tr_ood_auc_entropy}{std_tr_ood_auc_entropy}
                \plotwitherrorareas{data/entropy-vs-SPC-fashion_mnist-results-miniVGG-baseline-combined.csv}{magenta}{BL}{mean_tr_ood_auc_entropy}{std_tr_ood_auc_entropy}
                \plotwitherrorareas{data/entropy-vs-SPC-fashion_mnist-results-miniVGG-flipout-combined.csv}{gray}{VI}{mean_tr_ood_auc_entropy}{std_tr_ood_auc_entropy}
                \legend{}
            \end{axis}		            
        \end{tikzpicture}
        \caption{Train/OOD AUC Entropy}
    \end{subfigure}
    \begin{subfigure}{0.49\textwidth}
        \begin{tikzpicture}
            \begin{axis}[height = 0.18 \textheight, width = \textwidth, xlabel={Samples per Class}, ylabel={OOD AUC}, ymajorgrids=false, xmajorgrids=false, grid style=dashed, tick label style={font=\tiny,rotate=40}, xtick=data, xmode=log, log ticks with fixed point, ymin=0.2, ymax=1.0]
                
                \plotwitherrorareas{data/entropy-vs-SPC-fashion_mnist-results-miniVGG-deepensemble-combined.csv}{blue}{DE}{mean_tr_ood_auc_maxprob}{std_tr_ood_auc_maxprob}
                \plotwitherrorareas{data/entropy-vs-SPC-fashion_mnist-results-miniVGG-dropout-combined.csv}{red}{DO}{mean_tr_ood_auc_maxprob}{std_tr_ood_auc_maxprob}
                \plotwitherrorareas{data/entropy-vs-SPC-fashion_mnist-results-miniVGG-dropconnect-combined.csv}{green}{DC}{mean_tr_ood_auc_maxprob}{std_tr_ood_auc_maxprob}
                \plotwitherrorareas{data/entropy-vs-SPC-fashion_mnist-results-miniVGG-duq-combined.csv}{black}{DUQ}{mean_tr_ood_auc_maxprob}{std_tr_ood_auc_maxprob}
                \plotwitherrorareas{data/entropy-vs-SPC-fashion_mnist-results-miniVGG-baseline-combined.csv}{magenta}{BL}{mean_tr_ood_auc_maxprob}{std_tr_ood_auc_maxprob}
                \plotwitherrorareas{data/entropy-vs-SPC-fashion_mnist-results-miniVGG-flipout-combined.csv}{gray}{VI}{mean_tr_ood_auc_maxprob}{std_tr_ood_auc_maxprob}
                \plotwitherrorareas{data/maxprob-vs-SPC-fashion_mnist-results-miniVGG-gradient-l1_norm-combined.csv}{orange}{GD}{mean_tr_ood_auc_maxprob}{std_tr_ood_auc_maxprob}
                \legend{}
            \end{axis}		            
        \end{tikzpicture}
        \caption{Train/OOD AUC Max Prob}
    \end{subfigure}
    \caption{Comparison of Out of Distribution Performance on Fashion MNIST}
    \label{ood_comparison_fmnist}
\end{figure*}
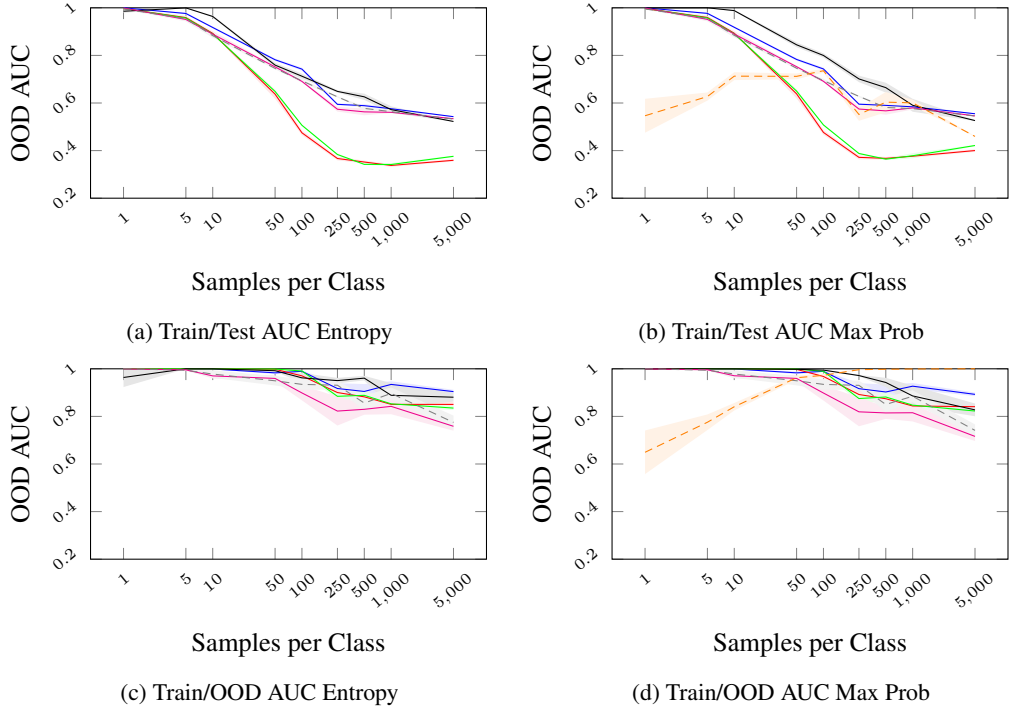
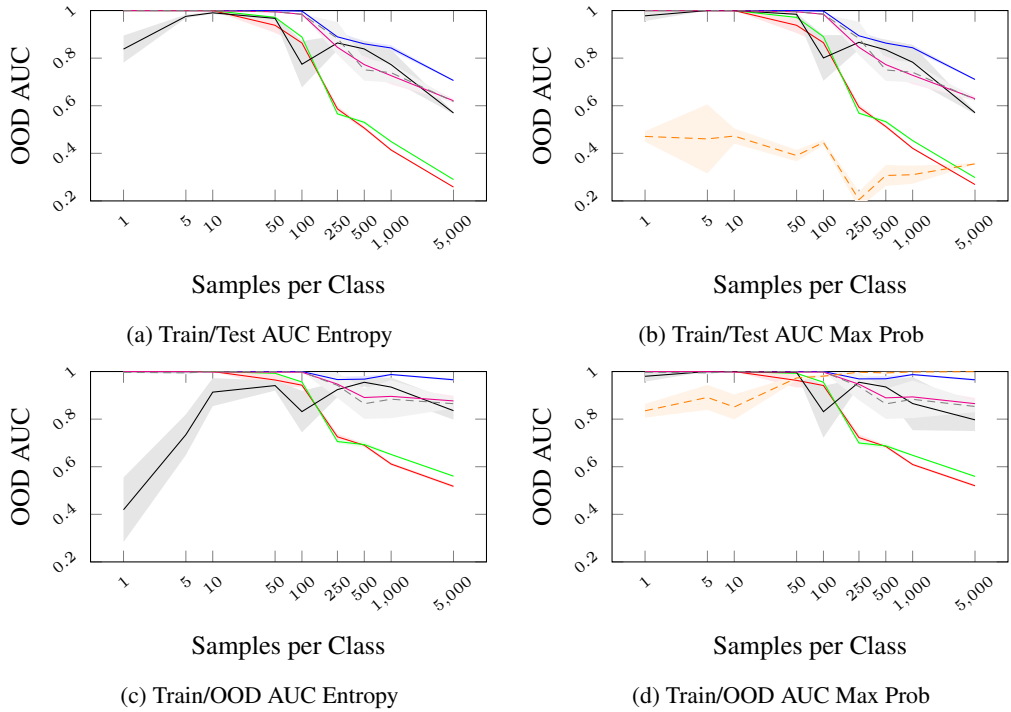
\begin{figure*}[!hb]
    \begin{subfigure}{0.49\textwidth}
        \begin{tikzpicture}
            \begin{axis}[height = 0.18 \textheight, width = \textwidth, xlabel={Samples per Class}, ylabel={OOD AUC}, ymajorgrids=false, xmajorgrids=false, grid style=dashed, tick label style={font=\tiny,rotate=40}, xtick=data, xmode=log, log ticks with fixed point, ymin=0.2, ymax=1.0]
                
                \plotwitherrorareas{data/entropy-vs-SPC-cifar10-results-miniVGG-deepensemble-combined.csv}{blue}{DE}{mean_tr_test_auc_entropy}{std_tr_test_auc_entropy}
                \plotwitherrorareas{data/entropy-vs-SPC-cifar10-results-miniVGG-dropout-combined.csv}{red}{DO}{mean_tr_test_auc_entropy}{std_tr_test_auc_entropy}
                \plotwitherrorareas{data/entropy-vs-SPC-cifar10-results-miniVGG-dropconnect-combined.csv}{green}{DC}{mean_tr_test_auc_entropy}{std_tr_test_auc_entropy}
                \plotwitherrorareas{data/entropy-vs-SPC-cifar10-results-miniVGG-duq-combined.csv}{black}{DUQ}{mean_tr_test_auc_entropy}{std_tr_test_auc_entropy}
                \plotwitherrorareas{data/entropy-vs-SPC-cifar10-results-miniVGG-baseline-combined.csv}{magenta}{BL}{mean_tr_test_auc_entropy}{std_tr_test_auc_entropy}
                \plotwitherrorareas{data/entropy-vs-SPC-cifar10-results-miniVGG-flipout-combined.csv}{gray}{VI}{mean_tr_test_auc_entropy}{std_tr_test_auc_entropy}
                \legend{}
            \end{axis}		            
        \end{tikzpicture}
        \caption{Train/Test AUC Entropy}
    \end{subfigure}    
    \begin{subfigure}{0.49\textwidth}
        \begin{tikzpicture}
            \begin{axis}[height = 0.18 \textheight, width = \textwidth, xlabel={Samples per Class}, ylabel={OOD AUC}, ymajorgrids=false, xmajorgrids=false, grid style=dashed, tick label style={font=\tiny,rotate=40}, xtick=data, xmode=log, log ticks with fixed point, ymin=0.2, ymax=1.0]
                
                \plotwitherrorareas{data/entropy-vs-SPC-cifar10-results-miniVGG-deepensemble-combined.csv}{blue}{DE}{mean_tr_test_auc_maxprob}{std_tr_test_auc_maxprob}
                \plotwitherrorareas{data/entropy-vs-SPC-cifar10-results-miniVGG-dropout-combined.csv}{red}{DO}{mean_tr_test_auc_maxprob}{std_tr_test_auc_maxprob}
                \plotwitherrorareas{data/entropy-vs-SPC-cifar10-results-miniVGG-dropconnect-combined.csv}{green}{DC}{mean_tr_test_auc_maxprob}{std_tr_test_auc_maxprob}
                \plotwitherrorareas{data/entropy-vs-SPC-cifar10-results-miniVGG-duq-combined.csv}{black}{DUQ}{mean_tr_test_auc_maxprob}{std_tr_test_auc_maxprob}
                \plotwitherrorareas{data/entropy-vs-SPC-cifar10-results-miniVGG-baseline-combined.csv}{magenta}{BL}{mean_tr_test_auc_maxprob}{std_tr_test_auc_maxprob}
                \plotwitherrorareas{data/entropy-vs-SPC-cifar10-results-miniVGG-flipout-combined.csv}{gray}{VI}{mean_tr_test_auc_maxprob}{std_tr_test_auc_maxprob}
                \plotwitherrorareas{data/maxprob-vs-SPC-cifar10-results-miniVGG-gradient-l1_norm-combined.csv}{orange}{GD}{mean_tr_test_auc_maxprob}{std_tr_test_auc_maxprob}
                \legend{}
            \end{axis}		            
        \end{tikzpicture}
        \caption{Train/Test AUC Max Prob}
    \end{subfigure}
    
    \begin{subfigure}{0.49\textwidth}
        \begin{tikzpicture}
            \begin{axis}[height = 0.18 \textheight, width = \textwidth, xlabel={Samples per Class}, ylabel={OOD AUC}, ymajorgrids=false, xmajorgrids=false, grid style=dashed, legend pos = south east, legend style={font=\scriptsize}, tick label style={font=\tiny,rotate=40}, xtick=data, xmode=log, log ticks with fixed point, ymin=0.2, ymax=1.0]
                
                \plotwitherrorareas{data/entropy-vs-SPC-cifar10-results-miniVGG-deepensemble-combined.csv}{blue}{DE}{mean_tr_ood_auc_entropy}{std_tr_ood_auc_entropy}  
                \plotwitherrorareas{data/entropy-vs-SPC-cifar10-results-miniVGG-dropout-combined.csv}{red}{DO}{mean_tr_ood_auc_entropy}{std_tr_ood_auc_entropy}
                \plotwitherrorareas{data/entropy-vs-SPC-cifar10-results-miniVGG-dropconnect-combined.csv}{green}{DC}{mean_tr_ood_auc_entropy}{std_tr_ood_auc_entropy}
                \plotwitherrorareas{data/entropy-vs-SPC-cifar10-results-miniVGG-duq-combined.csv}{black}{DUQ}{mean_tr_ood_auc_entropy}{std_tr_ood_auc_entropy}
                \plotwitherrorareas{data/entropy-vs-SPC-cifar10-results-miniVGG-baseline-combined.csv}{magenta}{BL}{mean_tr_ood_auc_entropy}{std_tr_ood_auc_entropy}
                \plotwitherrorareas{data/entropy-vs-SPC-cifar10-results-miniVGG-flipout-combined.csv}{gray}{VI}{mean_tr_ood_auc_entropy}{std_tr_ood_auc_entropy}
                \legend{}
            \end{axis}		            
        \end{tikzpicture}
        \caption{Train/OOD AUC Entropy}
    \end{subfigure}
    \begin{subfigure}{0.49\textwidth}
        \begin{tikzpicture}
            \begin{axis}[height = 0.18 \textheight, width = \textwidth, xlabel={Samples per Class}, ylabel={OOD AUC}, ymajorgrids=false, xmajorgrids=false, grid style=dashed, tick label style={font=\tiny,rotate=40}, xtick=data, xmode=log, log ticks with fixed point, ymin=0.2, ymax=1.0]
                
                \plotwitherrorareas{data/entropy-vs-SPC-cifar10-results-miniVGG-deepensemble-combined.csv}{blue}{DE}{mean_tr_ood_auc_maxprob}{std_tr_ood_auc_maxprob}
                \plotwitherrorareas{data/entropy-vs-SPC-cifar10-results-miniVGG-dropout-combined.csv}{red}{DO}{mean_tr_ood_auc_maxprob}{std_tr_ood_auc_maxprob}
                \plotwitherrorareas{data/entropy-vs-SPC-cifar10-results-miniVGG-dropconnect-combined.csv}{green}{DC}{mean_tr_ood_auc_maxprob}{std_tr_ood_auc_maxprob}
                \plotwitherrorareas{data/entropy-vs-SPC-cifar10-results-miniVGG-duq-combined.csv}{black}{DUQ}{mean_tr_ood_auc_maxprob}{std_tr_ood_auc_maxprob}
                \plotwitherrorareas{data/entropy-vs-SPC-cifar10-results-miniVGG-baseline-combined.csv}{magenta}{BL}{mean_tr_ood_auc_maxprob}{std_tr_ood_auc_maxprob}
                \plotwitherrorareas{data/entropy-vs-SPC-cifar10-results-miniVGG-flipout-combined.csv}{gray}{VI}{mean_tr_ood_auc_maxprob}{std_tr_ood_auc_maxprob}
                \plotwitherrorareas{data/maxprob-vs-SPC-cifar10-results-miniVGG-gradient-l1_norm-combined.csv}{orange}{GD}{mean_tr_ood_auc_maxprob}{std_tr_ood_auc_maxprob}
                \legend{}
            \end{axis}		            
        \end{tikzpicture}
        \caption{Train/OOD AUC Max Prob}
    \end{subfigure}
    \caption{Comparison of Out of Distribution Performance on CIFAR10}
    \label{ood_comparison_cifar10}
\end{figure*}

\FloatBarrier
\section{Gradient Uncertainty Performance}

\end{document}